\documentclass[10pt,logo,copyright]{nvidiatechreport}
\usepackage{subcaption}
\usepackage{wrapfig}
\usepackage{makecell}
\usepackage{multirow}

\usepackage[numbers, sort]{natbib}
\definecolor{citecolor}{HTML}{0071BC}
\definecolor{linkcolor}{HTML}{ED1C24}

\usepackage{nicefrac}

\usepackage[capitalize]{cleveref}
\crefname{section}{\S}{\S\S}
\crefname{subsection}{\S}{\S\S}
\crefname{table}{\text{Tab.}}{\text{Tab.}}
\Crefname{table}{Table}{Tables}
\crefname{figure}{\text{Fig.}}{\text{Fig.}}
\Crefname{figure}{Figure}{Figures}
\crefname{equation}{\text{Eq}}{\text{Eq}}

\newcommand{\ours}{Lyra 2.0\xspace}

\title{Lyra 2.0: Explorable Generative 3D Worlds}

\author{
  \textbf{Tianchang Shen}\textsuperscript{*} \quad
  \textbf{Sherwin Bahmani} \quad
  \textbf{Kai He} \quad
  \textbf{Sangeetha Grama Srinivasan} \quad
  \textbf{Tianshi Cao} \quad
  \textbf{Jiawei Ren} \newline
  \textbf{Ruilong Li} \quad
  \textbf{Zian Wang} \quad
  \textbf{Nicholas Sharp} \quad
  \textbf{Zan Gojcic} \quad
  \textbf{Sanja Fidler} \quad
  \textbf{Jiahui Huang} \quad
  \textbf{Huan Ling} \quad
  \textbf{Jun Gao} \newline
  \textbf{Xuanchi Ren}\textsuperscript{*} \\
  \small NVIDIA \\
  \small $^{*}$Equal contribution \\
  {\small \url{https://research.nvidia.com/labs/sil/lyra2/}}
}

\begin{document}
\maketitle
\vspace{-3 em}
\begin{figure}[ht!]
  \centering
  \includegraphics[width=0.9
  \linewidth]{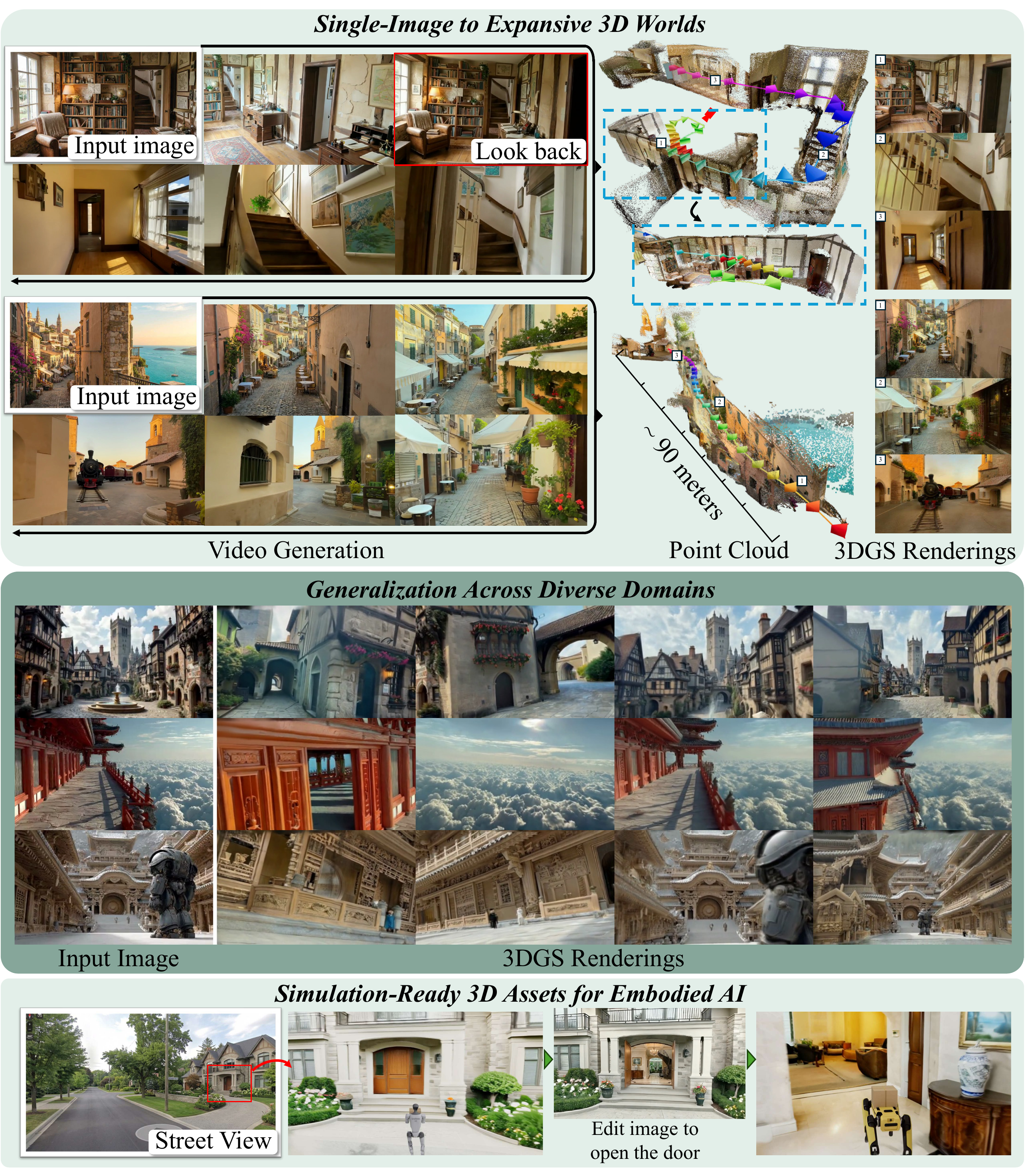}
  \vspace{-1 em}
  \caption{\textbf{\ours} enables long-horizon 3D-consistent scene generation from a single image. 
  Starting from an input image, users iteratively define camera motion to explore the scene, while \ours synthesizes spatially persistent video outputs that progressively expand the environment.
  These videos can be directly reconstructed into high-fidelity 3D Gaussians and surface meshes, yielding 3D assets deployable in simulation engines and interactive viewers.}
  \label{fig:teaser}
\end{figure}
\clearpage
\begin{abstract}
Recent advances in video generation enable a new paradigm for 3D
scene creation: generating camera-controlled videos that simulate scene
walkthroughs, then lifting them to 3D via feed-forward reconstruction techniques.
This \emph{generative reconstruction} approach combines the visual
fidelity and creative capacity of video models with 3D outputs ready for
real-time rendering and simulation. 
Scaling to large, complex environments requires 3D-consistent video generation over long camera trajectories with large viewpoint changes and location revisits, a setting where current video models degrade
quickly. 
Existing methods for long-horizon generation are fundamentally limited by two
forms of degradation: \emph{spatial forgetting} and \emph{temporal
drifting}. As exploration proceeds, previously observed regions fall
outside the model's temporal context, forcing the model to hallucinate
structures when revisited. Meanwhile, autoregressive generation
accumulates small synthesis errors over time, gradually distorting scene
appearance and geometry.
We present \ours, a framework for generating persistent, explorable 3D worlds at scale. 
To address spatial forgetting, we maintain per-frame 3D geometry and use it solely for information routing---retrieving relevant past frames and establishing dense correspondences with the target viewpoints---while relying on the generative prior for appearance synthesis.
To address temporal drifting, we train with self-augmented histories that expose the model to its own degraded outputs, teaching it to correct drift rather than propagate it.
Together, these enable substantially longer and 3D-consistent video trajectories, which we leverage to fine-tune feed-forward reconstruction models that reliably recover high-quality 3D scenes. 

\end{abstract}

\abscontent

\section{Introduction}
\label{sec:intro}

Trained on massive internet data, video diffusion models~\cite{wan2025wan,agarwal2025cosmos,Sora,veo} now exhibit remarkable visual fidelity and strong local 3D consistency between neighboring frames.
This progress enables \emph{generative reconstruction}~\cite{bahmani2025lyra}: given a single image and a prescribed camera trajectory, a video diffusion model synthesizes dense novel views that serve as virtual captures for feed-forward 3D reconstruction, recovering explicit scene geometry and appearance.
By replacing labor-intensive real-world capture with generative view synthesis, this paradigm enables scalable creation of diverse, high-quality, and even entirely imaginary 3D environments.

However, scaling generative reconstruction to large, complex environments---such as navigating across rooms or long city streets---requires maintaining 3D consistency over extended trajectories with substantial viewpoint changes and revisits.
Current video models generate frames autoregressively and struggle in such unbounded exploration scenarios, primarily suffering from two forms of degradation.
First, \emph{spatial forgetting}: as the camera moves, previously observed regions inevitably exceed the model's finite temporal context window. Upon revisiting these areas, the model is forced to hallucinate structures from scratch, breaking global layout consistency.
Second, \emph{temporal drifting}: autoregressive generation is inherently susceptible to error accumulation. Small per-step synthesis artifacts compound over time, leading to severe color shifts and structural distortions. This is further exacerbated during camera exploration, where continuously introduced unseen regions diminish visual overlap with early history frames, depriving the model of reliable geometric and texture constraints.

Recent efforts to mitigate spatial forgetting incorporate historical memory into the generation process.
A prominent line of work~\cite{ren2025gen3c, bahmani2025lyra, zhao2025spatia, zhou2025learning} maintains a cumulative 3D representation, conditioning subsequent frames on rendered views of the reconstructed geometry.
While providing explicit spatial constraints, this tightly coupled design suffers from error amplification: generative artifacts degrade the 3D geometry, which in turn produces flawed conditioning for future frames.
Alternatively, incorporating history frames directly into the context window via camera pose embeddings~\cite{cameractrl} avoids corrupted 3D intermediaries. Yet, this relies entirely on the model's self-attention to infer long-range geometric correspondences, which frequently fails under large viewpoint variations.
Instead, we bridge these two memory mechanisms by decoupling geometric tracking from pixel synthesis. We utilize an explicit 3D proxy solely for \emph{information routing}---retrieving relevant historical context and establishing spatial correspondences. Given this undistorted history context with dense spatial grounding, the actual novel view synthesis is left to the diffusion model's learned pixel prior, which resolves geometric inconsistencies and synthesizes novel views without propagating hard rendering artifacts.

To mitigate temporal drifting, existing strategies typically extend the temporal context length to anchor on past frames~\cite{zhang2025framepack}.
However, in scene exploration, camera motion inherently moves early frames out of the field of view, rendering long-context anchoring ineffective for suppressing drift in newly observed areas.
We propose alleviating the underlying training-inference discrepancy through a \emph{self-augmentation} training scheme.
By stochastically conditioning the network on its own one-step denoised predictions during training rather than perfect ground-truth frames, we expose the model to the exact error distributions encountered during autoregressive inference. Together with retrieving high-overlap history frames in the context window, the video model learns to actively mitigate drifting in recent generations with minimal computational overhead.

Equipped with these mechanisms, our model achieves highly persistent and long-horizon scene generation.
Nevertheless, videos synthesized by diffusion models inevitably contain minor multi-view inconsistencies that easily break traditional 3D reconstruction models, causing floaters and noisy artifacts.
To achieve reliable scene reconstruction, we employ a feed-forward 3D Gaussian Splatting (3DGS) pipeline~\cite{lin2025depth}. Fine-tuned on our generated sequences, this feed-forward model leverages its learned multi-view prior to tolerate minor inconsistencies, effectively bridging the domain gap and producing clean, coherent 3D structures.

We integrate these components into \ours, an interactive system for large-scale 3D scene exploration.
Starting from a single image, \ours empowers users to define arbitrary long-horizon camera trajectories and progressively reconstruct complex environments.
As demonstrated in \cref{fig:teaser}, our approach supports extensive scene navigation, including lookbacks and large-scale synthesis. The generated content can then be reliably reconstructed into high-quality 3D Gaussians and surface meshes with accurate geometry, ready for downstream applications in embodied AI and immersive rendering.

\section{Related Work}
\label{sec:related}

\noindent\textbf{Camera-Conditioned Video Generation.}
There have been significant advances in extending video diffusion models to incorporate camera control. Early approaches inject explicit camera parameterizations into the generative backbone. For instance, MotionCtrl~\cite{MotionCtrl} flattens per-frame camera pose matrices into vectors and injects them into intermediate feature representations of a pre-trained video diffusion model. Subsequent works~\cite{cameractrl, xu2024camco, bahmani2024ac3d, bahmani2024vd3d} adopt dense ray-based encodings using Pl\"ucker coordinates~\cite{chen2023ray,sitzmann2021light}, enabling pixel-wise camera conditioning and improved viewpoint control.
Following the success of Genie 3~\cite{ball2025genie}, an increasingly popular line of work~\cite{li2025hunyuan,mao2025yume,tang2025hunyuan,zhang2025matrix,he2025matrix} formulates camera control as an action-conditioning problem, where viewpoint changes are driven by discrete control signals such as keyboard inputs.
To further improve geometric faithfulness, recent approaches~\cite{yu2024viewcrafter,ren2025gen3c,wu2025video,yu2025trajectorycrafter,li2025magicworld,zhao2025spatia} introduce more structured 3D guidance signals beyond per-frame pose conditioning. These methods condition generation on renderings of estimated 3D geometry, such as global point cloud renderings or depth-warped images, to better constrain spatial structure during generation. GenWarp~\cite{seo2024genwarp} introduces correspondence-based conditioning but is limited to single-image diffusion models.

While these works produce compelling videos under viewpoint control, the underlying 3D consistency of the generated scenes is often limited and does not remain persistent when revisiting previously generated regions. Our work builds upon the camera-controlled video generation paradigm and addresses the fundamental problems of spatial forgetting and temporal drifting in long-horizon 3D-consistent generation.

\noindent\textbf{Memory-Aware Long Video Generation.}
Although camera conditioning enables controllable viewpoint changes, most video diffusion models remain constrained by a fixed temporal context window. As a result, long-horizon consistency degrades once previously generated content falls outside the attention span of the model. To address this limitation, recent works augment generative models with explicit memory mechanisms.
A first family of approaches~\cite{yu2025context,xiao2025worldmem,li2025vmem,gu2025long} relies on retrieval-based memory. These methods treat past frames as an external memory bank and dynamically select relevant observations to guide the next generation. For example, Context-as-Memory~\cite{yu2025context} and WorldMem~\cite{xiao2025worldmem} retrieve earlier frames based on field-of-view (FOV) overlap, while VMem~\cite{li2025vmem} performs geometry-aware retrieval using indexed 3D surface elements instead of purely view-based similarity.
A second line of work~\cite{zhou2025learning,liu2025dynamem,zhao2025spatia,wu2025geometry,li2025magicworld} enforces spatial persistence through explicit 3D representations accumulated over time. Rather than retrieving individual frames, these methods construct and maintain a global scene structure that serves as a unified memory for camera control and revisit consistency.
A third direction~\cite{po2025bagger,hong2025relic,savov2025statespacediffuser,zhang2025test,dalal2025one,hong2024slowfast} improves long-range temporal coherence by modifying the internal architecture of the generator, maintaining persistent latent states or key--value caches that propagate information across timesteps.
Orthogonally, FramePack~\cite{zhang2025framepack} compresses history frames into compact contextual slots through variable patchification based on temporal relevance, extending the effective context window without architectural changes.

In contrast to global 3D memory methods that rely on a single accumulated scene representation, we maintain per-frame 3D geometry and use it exclusively for information routing, \ie, retrieving relevant history frames and establishing dense geometric correspondences, while relying on the video model's generative prior for appearance synthesis. Combined with a self-augmentation strategy that mitigates temporal drifting, our approach enables scalable scene expansion and long-horizon 3D consistency under complex camera motion.

\noindent\textbf{3D Scene Generation.}
A prominent line of work~\cite{charatan2024pixelsplat, zhang2024gs, szymanowicz2024flash3d,ren2024scube,lin2025depth,lu2024infinicube} reconstructs 3D Gaussians~\cite{kerbl20233d} from one or multiple input views in a fully feed-forward manner.
Recent approaches combine generative modeling with feed-forward 3D reconstruction to reduce the reliance on densely sampled multi-view inputs. Bolt3D~\cite{szymanowicz2025bolt3d}, for example, trains a pointmap~\cite{dust3r} autoencoder to generate multi-view pointmaps, which are subsequently used for feed-forward 3D reconstruction. Wonderland~\cite{liang2024wonderland} utilizes a camera-controlled video diffusion model to synthesize multi-view imagery and then predicts 3D Gaussians with a dedicated feed-forward network. More recently, Lyra~\cite{bahmani2025lyra} adopts a camera-controlled video model as a teacher within a self-distillation framework, enabling the training of a student 3D reconstruction model without requiring real-world multi-view supervision. FlashWorld~\cite{li2026flashworld} further demonstrates efficient 3D scene generation using a distilled camera-controlled video diffusion model. WorldExplorer~\cite{schneider2025worldexplorer} generates navigable 3D scenes from text by iteratively producing camera-guided videos from panoramic initializations and fusing them into 3D Gaussians via per-scene optimization. Concurrently, Video-to-World~\cite{hoellein2026world} proposes a non-rigid alignment procedure to correct 3D inconsistencies in video generations before lifting into 3D. Free-Range Gaussians~\cite{shabanov2026free} tackles generative 3D reconstruction by using flow matching directly on the Gaussian parameters.

These methods achieve strong results but typically remain limited in view coverage. We instead generate long, 3D-consistent videos from a single image and lift them into large-scale 3D Gaussians and meshes via a scalable feed-forward reconstruction pipeline.

\section{Preliminaries}
\label{sec:preliminary}

\noindent\textbf{DiT-Based Latent Video Diffusion.}
Our method builds upon DiT-based latent video diffusion models~\cite{agarwal2025cosmos,wan2025wan}.
Given an RGB video of $F$ frames, $\mathbf{x} \in \mathbb{R}^{F \times 3 \times H \times W}$, a VAE encoder compresses it into a latent $\mathbf{z} = \mathcal{E}(\mathbf{x}) \in \mathbb{R}^{F' \times C \times h \times w}$, from which a decoder reconstructs $\hat{\mathbf{x}} = \mathcal{D}(\mathbf{z})$.
To jointly handle images and videos, modern causal video VAEs encode the first frame independently and temporally compress subsequent frames.
We adopt the Wan~2.1 VAE~\cite{wan2025wan}, which downsamples $8{\times}$ spatially and $4{\times}$ temporally, giving $F' = \lfloor(F{-}1)/4\rfloor + 1$, $C = 16$, $h = H/8$, $w = W/8$.
Generation is performed in this latent space via flow matching~\cite{lipman2022flow}: given a clean latent $\mathbf{z}_0$ and noise $\boldsymbol{\epsilon} \sim \mathcal{N}(0,\mathbf{I})$, we form $\mathbf{z}_t = (1{-}t)\,\mathbf{z}_0 + t\,\boldsymbol{\epsilon}$ for $t \in [0,1]$ and train a DiT $v_\theta$ to regress the velocity:
\begin{equation}
    \mathcal{L} = \mathbb{E}_{\mathbf{z}_0,\, t,\, \boldsymbol{\epsilon}} \left[\left\| v_\theta(\mathbf{z}_t, t, \mathbf{c}) - (\boldsymbol{\epsilon} - \mathbf{z}_0) \right\|^2 \right],
\end{equation}
where $\mathbf{c}$ denotes conditioning signals (\eg, text). Long videos can be produced by generating fixed-length segments autoregressively, conditioning each step on previously generated frames.

\noindent\textbf{Camera-Conditioned Video Generation.}
Generating 3D-consistent scene explorations requires the model to follow a prescribed camera trajectory.
For the $i$-th image, we denote the world-to-camera extrinsic as $\mathbf{T}_i = [\mathbf{R}_i \mid \mathbf{t}_i] \in \mathbb{R}^{3 \times 4}$, intrinsic $\mathbf{K}_i \in \mathbb{R}^{3 \times 3}$, and estimated depth map $D_i \in \mathbb{R}^{H \times W}$~\cite{ren2025gen3c}.
Two complementary strategies exist for injecting camera information into a DiT.
\textbf{Depth-based warping}~\cite{ren2025gen3c} forward-warps the most recent frame $I_j$ to each target viewpoint $(\mathbf{T}_i, \mathbf{K}_i)$ using its depth $D_j$, encodes and concatenates the renderings with the denoising latent.
We find that within Wan~2.1~\cite{wan2025wan}, this mechanism alone already delivers accurate camera control even along long trajectories.
However, when the viewpoint change is large enough that no warped pixels land on the target view, the control signal is lost entirely, and the visual quality degrades significantly.
We therefore complement it with \textbf{Pl\"ucker ray injection}~\cite{cameractrl}, which computes 6D ray coordinates $\mathbf{r}_i(u,v) = (\mathbf{d},\; \mathbf{o} \times \mathbf{d}) \in \mathbb{R}^6$ per pixel, projects them to the DiT's hidden dimension via an MLP, and adds them to token features, providing an extra hint in case of drastic viewpoint changes. 

\noindent\textbf{Context Compression via FramePack.}
We adopt FramePack~\cite{zhang2025framepack} to compress the history context and mitigate drifting. We describe its details here and discuss additional strategies to further reduce drifting in~\cref{sec:temporal_memory}. 
FramePack compresses history frames with variable patchification kernels by temporal proximity: recent frames use a small kernel for fine-grained tokenization, while distant frames use a large kernel for aggressive compression. This allows the model to attend to a long temporal horizon within a fixed token budget. Typically, the temporal history is organized as:
\begin{equation}
    \underbrace{\texttt{f$1$k$1$}}_{\text{anchor}} \;\;
    \underbrace{\texttt{f$16$k$4$} \;\; \texttt{f$2$k$2$} \;\; \texttt{f$1$k$1$}}_{\text{temporal slots}} \;\;
    \underbrace{\texttt{g$20$}}_{\text{generate}},
\label{eq:framepack}
\end{equation}
where \texttt{f$n$k$m$} denotes $n$ frames compressed with spatial subsampling factor $m$, and \texttt{g$20$} is the 20-frame generation target. The anchor frame (the initial image $I_0$) is always included at full resolution as a fixed reference point, serving as an early-established endpoint~\cite{zhang2025framepack} that prevents the model from drifting away from the original scene appearance.

\section{Method}
\label{sec:method}

\subsection{Overview}
\label{sec:overview}
Given a single input image $I_0$ and a camera trajectory $\{(\mathbf{T}_i, \mathbf{K}_i)\}_{i=0}^{T-1}$, our goal is to generate a long, camera-controlled video that maintains global 3D consistency across all frames and can be lifted into an explorable 3D scene.

\begin{figure*}[t]
  \centering
  \includegraphics[width=\linewidth]{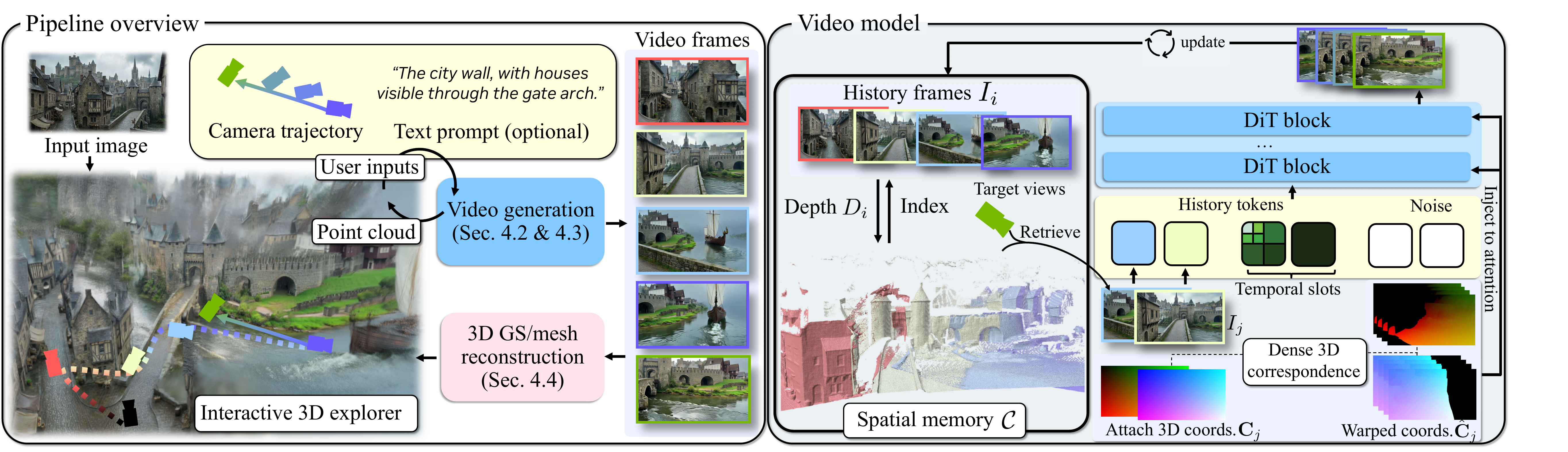}
  \caption{\textbf{Method overview.} \textbf{(Left)} Given an input image, \ours iteratively generates video segments guided by a user-defined camera trajectory from an interactive 3D explorer and an optional text prompt, lifting each segment into 3D point clouds fed back for continued navigation. Generated video frames are finally reconstructed and exported as 3D Gaussians or meshes. \textbf{(Right)} At each step, history frames with maximal visibility of the target views are retrieved from the spatial memory. Their canonical coordinates are warped to establish dense 3D correspondences and injected into DiT via attention, together with compressed temporal history.} %
  \label{fig:method}
\end{figure*}

As illustrated in~\cref{fig:method}, \ours generates long videos through an autoregressive \textbf{retrieve--generate--update} loop. At each iteration, the user first provides a 3D camera trajectory and an optional text prompt to guide outpainting. Then we (i)~\emph{retrieve} history frames whose 3D content is most relevant to the target viewpoint, (ii)~\emph{generate} the next video segment conditioned on both temporal history and retrieved spatial context, and (iii)~\emph{update} the memory with the newly generated frames. At the core of this pipeline are two mechanisms that address the key challenges in long-horizon autoregressive generation: \emph{anti-forgetting} (\cref{sec:persistent_video}), which builds a spatial memory with per-frame 3D geometry and bridges it with the video model's context to maintain spatial consistency when revisiting previously explored regions, and \emph{anti-drifting} (\cref{sec:temporal_memory}), which adaptively compresses history frames and mitigates quality degradation over long sequences.
The memory grows with each iteration step, enabling the model to maintain consistency over arbitrarily long trajectories and across revisits to previously explored regions.
The generated long video is then lifted into explicit 3D representations via feed-forward 3D reconstruction (\cref{sec:gs_reconstruction}).

\subsection{Anti-Forgetting for 3D-Persistent Video Generation}
\label{sec:persistent_video}

Achieving long-range spatial consistency requires recalling geometrically relevant history observations regardless of their temporal distance.
Our core intuition is to use noisy 3D geometry estimation exclusively for \emph{information routing}---selecting which history observations are relevant and establishing geometric correspondence between history and future viewpoints---while the video model handles all appearance synthesis and resolves inconsistencies between observations.
Following this intuition, we first build a \emph{3D cache} that stores per-frame geometry information, and design a \emph{retrieval strategy} that selects the most informative history frames for a given target viewpoint to condition the video model.

\noindent\textbf{Building the 3D Cache.}
We maintain a 3D cache $\mathcal{C}$ that grows incrementally as the video is generated. For each frame $I_i$ with estimated depth $D_i$~\cite{lin2025depth} and camera intrinsic and extrinsic $(\mathbf{T}_i, \mathbf{K}_i)$, our 3D cache maintains two components: (i)~the \emph{full-resolution depth map} $D_i$ and camera parameters;
(ii)~a \emph{downsampled point cloud} $\mathbf{P}_i \in \mathbb{R}^{(H/d) \times (W/d) \times 3}$, obtained by subsampling the depth map by factor $d$ and unprojecting it into world coordinates.
This first component preserves full geometric precision for correspondence computation, while the second one is exclusively used for efficient retrieval. 

Critically, the cache stores the geometry of each frame independently, and we never fuse them into a single global point cloud. This is particularly important in long-horizon generation, where depth estimation quality inevitably degrades over time, since it runs on the generated frames rather than real images. By maintaining per-frame point clouds, we avoid accumulating cross-view misalignment from imperfect depth into a single corrupted reconstruction.

\noindent\textbf{Geometry-Aware Retrieval.}
Since the context window of a video model is limited, selecting the most informative history frames is critical for maximizing long-range consistency and efficiency. At each autoregressive step, we select $N_s$ history frames whose 3D content is most visible from the target viewpoint. 
To achieve this, we compute the \emph{visibility score} $\phi$ of each history frame. 
Specifically, given a target camera $(\mathbf{T}^*, \mathbf{K}^*)$, we project every downsampled point cloud $\mathbf{P}_i$ onto the target image plane. Then, for each pixel on the target image plane, we compute the minimum projected depth over all frames to handle occlusion. A point is considered visible if and only if the difference between its depth and the minimum depth is less than a threshold $\delta$. The visibility score $\phi(i)$ of frame $i$ is the count of its visible points.
During training, we sample the history frames proportional to visibility scores $\phi(i)$ to make the model robust to different frame retrieval results.
At inference, we greedily maximize coverage: iteratively selecting the frame that covers the most not-yet-covered target pixels, up to $N_s$ frames. This avoids redundant selection of nearby viewpoints and maximizes the collective spatial coverage. 

With this mechanism, even when the camera revisits a region hundreds of frames later---far beyond the model's temporal context window---our retrieval can naturally recall the relevant observations via their 3D overlap.

\noindent\textbf{Injecting Spatial Memory into the Video Model.}
Having retrieved the most relevant history frames $\{I_j\}_{j=0}^{N_s-1}$, we inject them into the video model as \emph{spatial slots}: each retrieved frame is encoded independently by the VAE as image tokens (without temporal compression) and placed alongside the temporal FramePack slots and generation tokens. We apply the same variable-kernel spatial compression from FramePack to both the temporal and spatial slots. The full context layout is:
\begin{equation*}
    \underbrace{\texttt{f1k1}}_{\text{anchor}} \;\;
    \underbrace{\texttt{f4k2} \;\; \texttt{f1k1}}_{\text{spatial slots}} \;\;
    \underbrace{\texttt{f16k4} \;\; \texttt{f2k2} \;\; \texttt{f1k1}}_{\text{temporal slots}} \;\;
    \underbrace{\texttt{g20}}_{\text{generate}},
\end{equation*}
where spatial slots contribute $N_s{=}5$ retrieved frames: 4 frames at subsampling factor 2 and 1 frame at full resolution. All tokens are jointly processed by the full DiT self-attention.

While prior retrieval-based approaches~\cite{yu2025context,xiao2025worldmem} inject history frames in a similar fashion, they lack geometric grounding for precise multi-view alignment. To address this, we further establish dense correspondences via \emph{canonical coordinate warping}: for the $j$-th retrieved frame, we assign a canonical coordinate map $\mathbf{C}_j \in [-1,1]^{3 \times H \times W}$ whose three channels are $(u, v, 2 \cdot\frac{j}{N_s} - 1)$ where $(u, v)$ encodes the normalized spatial position. We then forward-warp $\mathbf{C}_j$ using the full-resolution depth:
\begin{equation}
    \hat{\mathbf{C}}_j = \texttt{FwdWarp}(\mathbf{C}_j,\; D_{s_j},\; \mathbf{T}_{s_j},\; \mathbf{T}^*,\; \mathbf{K}_{s_j},\; \mathbf{K}^*).
\end{equation}
We additionally warp the depth as a fourth channel, yielding a 4-channel map $[\hat{\mathbf{C}}_j;\, \hat{D}_j]$ per retrieved frame. When fewer than $N_s$ frames are retrieved, missing slots are padded so the model can distinguish real correspondences from empty slots. 
To feed the warped correspondence maps into DiT, we encode them via positional encoding and aggregate through a learned MLP. The output embeddings are added to the tokens at the self-attention layer of every transformer block.

Note that we warp canonical coordinates rather than RGB images for a specific reason: warped RGB inevitably contains disocclusion holes, stretching artifacts, and depth-boundary bleeding. If conditioned on such images, the video model tends to re-generate these artifacts---the warped image acts as a crutch that bypasses the generative prior rather than informing it. Canonical coordinates carry the same geometric correspondence information without any appearance content, leaving appearance synthesis entirely to the video model.

In summary, our video model context comprises three complementary signals: (1) the retrieved history frames $\{I_j\}_{j=0}^{N_s-1}$ encoded as spatial slots; (2) the forward-warped correspondence maps $[\hat{\mathbf{C}}_j;\, \hat{D}_j]_{j=0}^{N_s-1}$; and (3) the compressed temporal history via FramePack (\cref{eq:framepack}).

\subsection{Anti-Drifting for Long-Horizon Video Generation}
\label{sec:temporal_memory}
The root cause of drifting is \emph{observation bias}~\cite{zhang2025framepack}: during training, the model conditions on ground-truth history frames, but at inference it must condition on its own imperfect outputs. This train-test discrepancy means that per-step errors---color shifts, blurring, distortions---go uncorrected and compound across autoregressive steps, gradually degrading quality.
Context compression via FramePack (\cref{sec:preliminary}) alleviates drift by extending the temporal horizon and anchoring generation to the original image, but it does not close the fundamental observation bias gap. We therefore complement it with a \emph{self-augmentation} training strategy that directly reduces the train-test discrepancy.

\noindent\textbf{Self-Augmentation Training.}
Related approaches such as Self-Forcing~\cite{huang2025selfforcing} mitigate drifting by conditioning the model on its own predictions during training, but are primarily designed for causal network architectures. 
Directly applying self-forcing to our bi-directional video model is prohibitively expensive: each history segment would require full bi-directional attention and multi-step denoising (e.g., 35 steps) to simulate the model's inference-time outputs.

To address this, we introduce a lightweight \emph{self-augmentation} strategy. Consider an autoregressive training step with ground-truth history frames $\mathbf{x}^{\text{hist}}$ and current chunk frames $\mathbf{x}^{\text{cur}}$. Since our VAE is causal, encoding the current chunk depends on the temporal cache from the history segment. We encode both using clean ground-truth frames: $\mathbf{z}_0^{\text{hist}} = \mathcal{E}(\mathbf{x}^{\text{hist}})$ and $\mathbf{z}_0^{\text{cur}} = \mathcal{E}(\mathbf{x}^{\text{cur}} \mid \mathbf{x}^{\text{hist}})$, where the conditioning notation denotes the causal VAE cache dependency.

With probability $p_{\text{aug}}$, we corrupt the history latent by sampling $t \sim \mathcal{U}(0, 0.5)$ and adding noise according to the flow matching schedule:
\begin{equation}
    \mathbf{z}_t^{\text{hist}} = (1 - t)\,\mathbf{z}_0^{\text{hist}} + t\,\boldsymbol{\epsilon}, \quad \boldsymbol{\epsilon} \sim \mathcal{N}(0, \mathbf{I}).
\end{equation}
The video model then performs one-step denoising to produce an approximate reconstruction:
\begin{equation}
    \tilde{\mathbf{z}}_0^{\text{hist}} = \mathbf{z}_t^{\text{hist}} - t \cdot v_\theta(\mathbf{z}_t^{\text{hist}}, t, \mathbf{c}),
\end{equation}
and we replace $\mathbf{z}_0^{\text{hist}}$ with $\tilde{\mathbf{z}}_0^{\text{hist}}$ as the DiT's conditioning context. Crucially, the target latent $\mathbf{z}_0^{\text{cur}}$ is always encoded with the \emph{clean} history cache, and the flow matching loss supervises the DiT to denoise toward this clean $\mathbf{z}_0^{\text{cur}}$ despite receiving corrupted conditioning. This teaches the model to recover high-quality outputs from imperfect history context, effectively learning to counteract drifting artifacts during autoregressive inference. The overall overhead is minimal, requiring only one additional DiT forward pass.

\subsection{3D Reconstruction}
\label{sec:gs_reconstruction}
We lift the generated videos from \ours into explicit 3D representations for downstream applications such as embodied AI simulation and virtual reality.

\noindent\textbf{3D Gaussian Splatting.}
We adopt Depth Anything v3 (DAv3)~\cite{lin2025depth}, a feed-forward 3D foundation model that predicts per-pixel 3DGS attributes from input images. However, the pretrained DAv3 model exhibits two main limitations in our setting.
\emph{First}, DAv3 predicts one Gaussian per pixel, which leads to an excessively large number of Gaussians for the high-resolution inputs produced by our system. To address this, we modify the Gaussian DPT head in the DAv3 architecture to produce a feature map downsampled by a factor of $k \times k$. This allows the network to process the original high-resolution images while reducing the number of predicted Gaussians by $k^2$, yielding a more compact representation suitable for real-time rendering and data streaming.
\emph{Second}, DAv3 is not optimized for generated data, where small geometric inconsistencies are common. To improve robustness, we fine-tune the model on scenes generated by our video model. Similar to Lyra~\cite{bahmani2025lyra}, this improves robustness to artifacts commonly present in generative data.

\noindent\textbf{Surface Mesh Extraction.}
After obtaining the 3DGS, we further extract a surface mesh.
Specifically, we develop a hierarchical sparse grid approach for large-scale mesh extraction based on OpenVDB~\cite{museth2013vdb,williams2024fvdb}, allocating fine grid cells near the generation viewpoints and coarse cells in the distant background. The median depth from the Gaussian reconstruction is rasterized as a depth map in each view with normals computed as the gradient of depth, and we use the resulting oriented point cloud to construct a signed distance function on the sparse grid. Surfaces are extracted via marching cubes, stitched across hierarchy levels, and decimated for efficient downstream processing.

\subsection{Distilled Model for Accelerated Inference}
\label{sec:dmd}

We additionally train a distilled version of our model using Distribution Matching Distillation (DMD)~\cite{yin2024dmd} to accelerate inference. Starting from our trained teacher model, we distill a student model that generates videos in 4 denoising steps instead of 35. We also distill the classifier-free guidance into the student, eliminating the need for separate conditional and unconditional forward passes at inference. During distillation, we retain our self-augmentation strategy so that the student remains robust to autoregressive error accumulation. Combined, the reduced sampling steps and single-pass inference reduce the per-step generation time by roughly $13{\times}$ while maintaining comparable visual quality for interactive use cases.

\section{Experiments}
\label{sec:experiments}

\subsection{Training Details}
\label{sec:exp_setup}

\begin{table*}[tb]
  \caption{Quantitative comparison on single-view to long video generation. Best results are shown in \textbf{bold} and second best are \underline{underlined}.
  }
  \label{tab:long_video}
  \centering
  \resizebox{\linewidth}{!}{
  \begin{tabular}{@{}lcccccccccccccc@{}}
    \toprule
    \multirow{2}{*}[-3pt]{Method} & \multicolumn{7}{c}{DL3DV} & \multicolumn{7}{c}{Tanks-and-Temples} \\
    \cmidrule(lr){2-8} \cmidrule(l){9-15}
    & SSIM$\uparrow$ & LPIPS$\downarrow$ & FID$\downarrow$ & \makecell{Subjective\\Qual.$\uparrow$} & \makecell{Style\\Consist.$\uparrow$} & \makecell{Camera\\Ctrl.$\uparrow$} & \makecell{Reproj.\\Err.$\downarrow$}
    & SSIM$\uparrow$ & LPIPS$\downarrow$ & FID$\downarrow$ & \makecell{Subjective\\Qual.$\uparrow$} & \makecell{Style\\Consist.$\uparrow$} & \makecell{Camera\\Ctrl.$\uparrow$} & \makecell{Reproj.\\Err.$\downarrow$} \\
    \midrule
    GEN3C~\cite{ren2025gen3c}   & 0.346 & 0.535 & 58.96 & 24.60 & 76.77 & \textbf{69.54} & \textbf{0.068} & 0.350 & 0.589 & 79.07 & 21.75 & 75.54 & \textbf{70.91} & \textbf{0.054} \\
    Yume1.5~\cite{mao2025yume} & 0.342 & 0.719 & 84.84 & 22.80 & 66.73 & -- & 0.095 & 0.348 & 0.702 & 89.69 & 28.68 & 78.63 & -- & 0.083 \\
    CaM~\cite{yu2025context}   & 0.370 & 0.562 & 50.43 & 35.19 & 82.63 & 42.71 & \underline{0.069} & 0.367 & 0.605 & 59.20 & 34.22 & \underline{82.83} & 31.86 & \underline{0.056} \\
    VMem~\cite{li2025vmem}     & 0.331 & 0.744 & 120.59 & 18.54 & 76.14 & 0.68 & 0.268 & 0.338 & 0.767 & 136.48 & 16.21 & 70.54 & 0.00 & 0.263 \\
    SPMem~\cite{wu2025video}   & \underline{0.383} & 0.522 & 53.77 & 38.32 & 82.79 & 62.05 & 0.074 & \underline{0.383} & 0.571 & 60.11 & \underline{34.41} & 79.68 & 45.07 & 0.059 \\
    HY-WorldPlay~\cite{hyworld2025} & 0.373 & 0.765 & 139.36 & 4.79 & 54.62 & -- & 0.092 & 0.380 & 0.796 & 163.54 & 3.24 & 48.22 & -- & 0.084 \\
    \midrule
    \textbf{Ours}  & \textbf{0.388} & \textbf{0.498} & \textbf{43.43} & \underline{44.54} & \underline{87.46} & 64.67 & 0.076 & \textbf{0.384} & \underline{0.552} & \underline{51.33} & \textbf{43.35} & \textbf{85.07} & \underline{63.87} & 0.069 \\
    \textbf{Ours DMD}  & 0.359 & \underline{0.507} & \underline{43.63} & \textbf{45.21} & \textbf{88.57} & \underline{65.64} & 0.088 & 0.362 & \textbf{0.545} & \textbf{49.71} & \underline{43.02} & 78.91 & 58.12 & 0.077 \\
    \bottomrule
  \end{tabular}}
\end{table*}

\noindent\textbf{Datasets.}
We train our model on DL3DV~\cite{ling2024dl3dv}, which contains 10K long video clips of diverse real-world scenes. We estimate camera poses using ViPE~\cite{huang2025vipe} and predict per-frame depth with Depth Anything V3~\cite{lin2025depth}. Video captions are generated using Qwen3-VL-8B-Instruct~\cite{bai2025qwen3vl}.

\noindent\textbf{Paired Data Curation.}
For real-world videos from DL3DV, we sample $1{,}000$ frames per video.
During training, we construct conditioning--target pairs using two complementary strategies.
With 30\% probability, we train in image-to-video (I2V) mode, where the model generates the first $L = 80$ consecutive frames conditioned on a single initial frame.
With the remaining 70\% probability, we perform autoregressive chunk-based training.
Specifically, given a sequence of $T$ frames, we uniformly sample a segment index $s \in [0, S_{\max})$, where $S_{\max} = \lfloor (T - 1) / L \rfloor - 1$.
The history window spans frames $[0, \, s \cdot L + 1)$ as conditioning context, and the ground-truth target consists of the next $L$ consecutive frames in segment $s+1$.

\subsection{Evaluation on Long Video Generation}
\label{sec:exp_long_video}

\begin{figure*}[t]
  \centering
  \includegraphics[width=\linewidth]{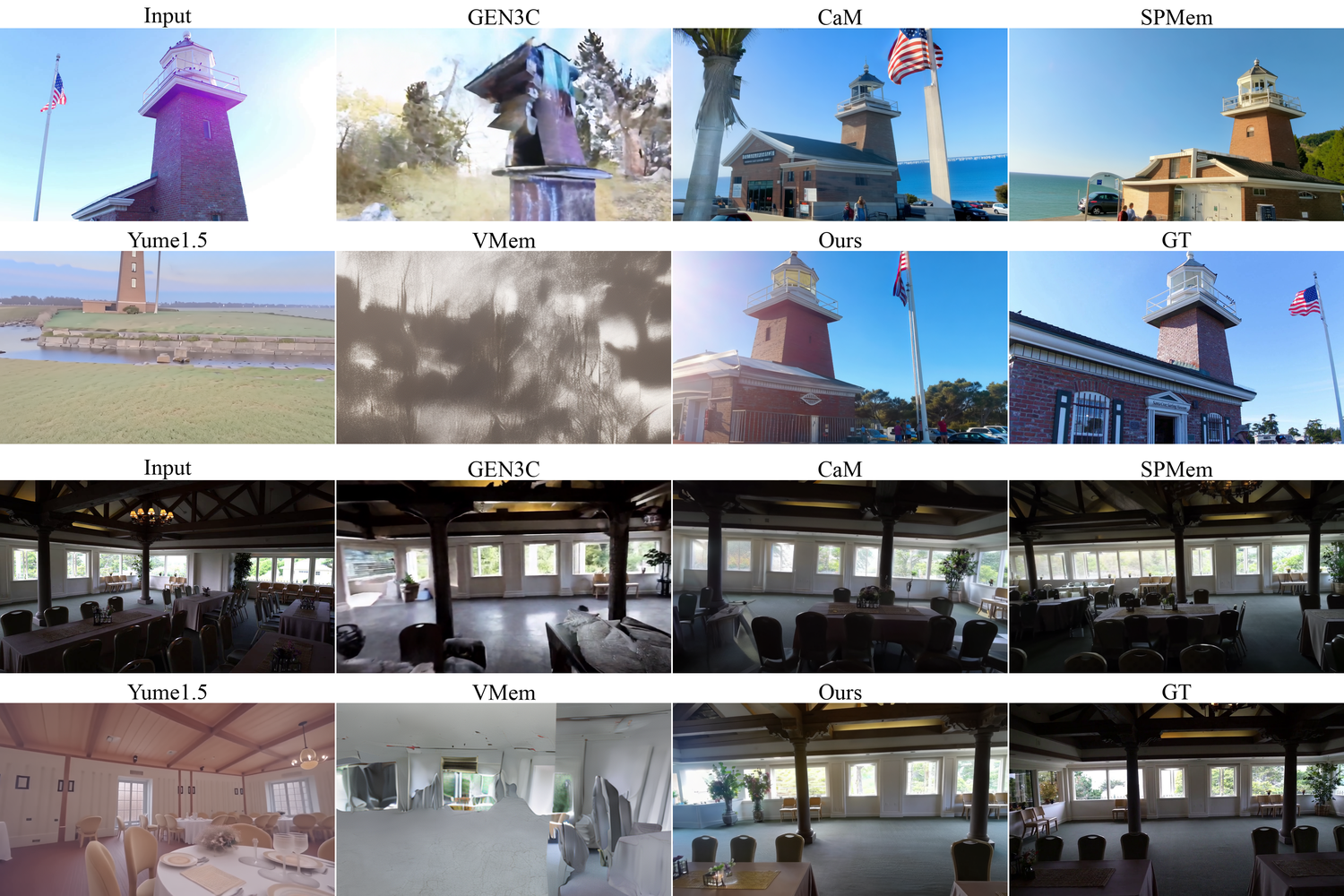}
  \caption{\textbf{Video generation comparisons.} Given a single input image from Tanks and Temples, we compare long-horizon generations (${\sim}$frame 800+) from all evaluated video models. Baselines exhibit severe quality degradation, geometric distortions, or content drifting at long horizons, while our method maintains realistic structures and appearances.}
  \label{fig:comparisons}
\end{figure*}

\noindent\textbf{Baselines and Metrics.}
We compare against recent camera-controllable long video generation methods with memory mechanisms: Yume-1.5~\cite{mao2025yume}, GEN3C~\cite{ren2025gen3c}, Context as Memory (CaM)~\cite{yu2025context}, VMem~\cite{li2025vmem}, SPMem~\cite{wu2025video}, and concurrent work HY-WorldPlay~\cite{hyworld2025}. Yume-1.5 is a FramePack-based method that relies solely on temporal context without spatial memory. GEN3C, CaM, VMem, and SPMem condition generation on multi-view history frames to maintain memory consistency. SPMem accumulates history frames into a global point cloud for conditioning. HY-WorldPlay uses discrete action control (keyboard inputs) rather than explicit camera trajectory conditioning. Since CaM and SPMem are not open-sourced, we re-implement them based on Wan2.1-14B~\cite{wan2025wan}.

All methods are evaluated on DL3DV-Evaluation~\cite{ling2024dl3dv} for in-domain testing and Tanks and Temples~\cite{knapitsch2017tanks} for out-of-domain generalization. We follow standard protocol~\cite{ren2022look,ren2025gen3c,yu2025context} and report SSIM, LPIPS, and Fr\'{e}chet Inception Distance (FID). Since standard metrics are insufficient for evaluating long video generation, we additionally adopt metrics from WorldScore~\cite{duan2025worldscore}: Subjective Quality Score for human perceptual quality, Style Consistency Score to detect visual drifting between the first and last frames, and Camera Controllability Score to measure camera pose accuracy. We further report reprojection error, computed by estimating per-frame depth with an off-the-shelf SLAM system~\cite{huang2025vipe}, to verify 3D consistency of the generated videos.

\noindent\textbf{Quantitative Comparison.}
As shown in Tab.~\ref{tab:long_video}, our method achieves the best results on both datasets across nearly all metrics, validating our anti-forgetting and anti-drifting mechanisms: 3D geometry serves as an information routing signal to enforce long-range consistency without sacrificing generation quality, while context compression and self-augmentation prevent quality degradation over long horizons.
Among the baselines, each addresses only one aspect of this challenge. GEN3C~\cite{ren2025gen3c} achieves the best Camera Controllability and Reprojection Error through explicit depth-warped conditioning, but this rigid geometric constraint degrades generation quality, as reflected by its low Subjective Quality and SSIM. CaM~\cite{yu2025context} and SPMem~\cite{wu2025video} are the strongest competitors on quality metrics thanks to their multi-view history memory, but their implicit camera conditioning leads to substantially lower Camera Controllability. SPMem's global point cloud conditioning also introduces geometric errors over long horizons, resulting in more pronounced drifting as reflected by lower Style Consistency. VMem~\cite{li2025vmem} struggles to maintain coherence over long horizons, resulting in the weakest scores across nearly all metrics. Yume-1.5~\cite{mao2025yume} and HY-WorldPlay~\cite{hyworld2025} lack explicit camera trajectory conditioning, failing to follow the specified viewpoints; HY-WorldPlay further suffers from severe temporal drifting, leading to substantial quality degradation.
In contrast, our framework bridges this gap, achieving both high visual fidelity and accurate camera control simultaneously.

\begin{table*}[tb]
  \caption{Quantitative comparison on 3D scene generation.
    Best results are shown in \textbf{bold} and second best are \underline{underlined}.
  }
  \label{tab:scene_gen}
  \centering
  \resizebox{\linewidth}{!}{
  \begin{tabular}{@{}lcccccccc@{}}
    \toprule
    \multirow{2}{*}{Method} & \multicolumn{4}{c}{DL3DV} & \multicolumn{4}{c}{Tanks-and-Temples} \\
    \cmidrule(lr){2-5} \cmidrule(lr){6-9}
    & LPIPS-P$\downarrow$ & LPIPS-G$\downarrow$ & FID$\downarrow$ & Subj. Qual.$\uparrow$
    & LPIPS-P$\downarrow$ & LPIPS-G$\downarrow$ & FID$\downarrow$ & Subj. Qual.$\uparrow$ \\
    \midrule
    GEN3C~\cite{ren2025gen3c} + DAv3   & 0.504 & 0.649 & 99.83  & 11.00 & 0.511 & 0.694 & 125.19 & 5.38  \\
    Yume1.5~\cite{mao2025yume} + DAv3  & 0.598 & 0.806 & 121.61 & 0.22  & 0.575 & 0.794 & 113.25 & 0.79  \\
    CaM~\cite{yu2025context} + DAv3    & 0.433 & 0.668 & 94.04  & 12.16 & 0.423 & 0.693 & 94.02  & 9.79  \\
    VMem~\cite{li2025vmem} + DAv3      & 0.593 & 0.836 & 206.88 & 2.00  & 0.597 & 0.832 & 211.72 & 3.76  \\
    SPMem~\cite{wu2025video} + DAv3    & 0.419 & 0.625 & 93.56  & 13.72 & 0.412 & 0.666 & 94.11  & 9.95  \\
    \midrule
    Ours + DAv3                        & \underline{0.413} & \underline{0.603} & \underline{74.39}  & \underline{17.02} & \underline{0.409} & \underline{0.648} & \underline{79.36}  & \underline{14.42} \\
    \textbf{Ours Full}                 & \textbf{0.381} & \textbf{0.579} & \textbf{65.94} & \textbf{20.52} & \textbf{0.372} & \textbf{0.629} & \textbf{72.47} & \textbf{18.80} \\
    \bottomrule
  \end{tabular}
  }
\end{table*}

\noindent\textbf{Qualitative Comparison.}
In Fig.~\ref{fig:comparisons}, we visualize single-image to long-video generation results. The shown images correspond to approximately frame 800, illustrating the challenges baselines face in maintaining realistic content over long generation horizons. VMem exhibits severe quality degradation and structural collapse; GEN3C and Yume-1.5 suffer from geometric distortions; CaM and SPMem maintain reasonable quality but show noticeable drifting. In contrast, our method maintains realistic geometric structures and appearances with respect to the input when revisiting regions.

\noindent\textbf{Distilled Model.}
As shown in Tab.~\ref{tab:long_video}, our DMD-distilled model (4 steps) achieves comparable or even slightly better per-frame quality (LPIPS, FID) compared to the full model (35 steps), while camera controllability decreases moderately due to the reduced number of denoising steps.

\begin{figure*}[t]
  \centering
  \includegraphics[width=\linewidth]{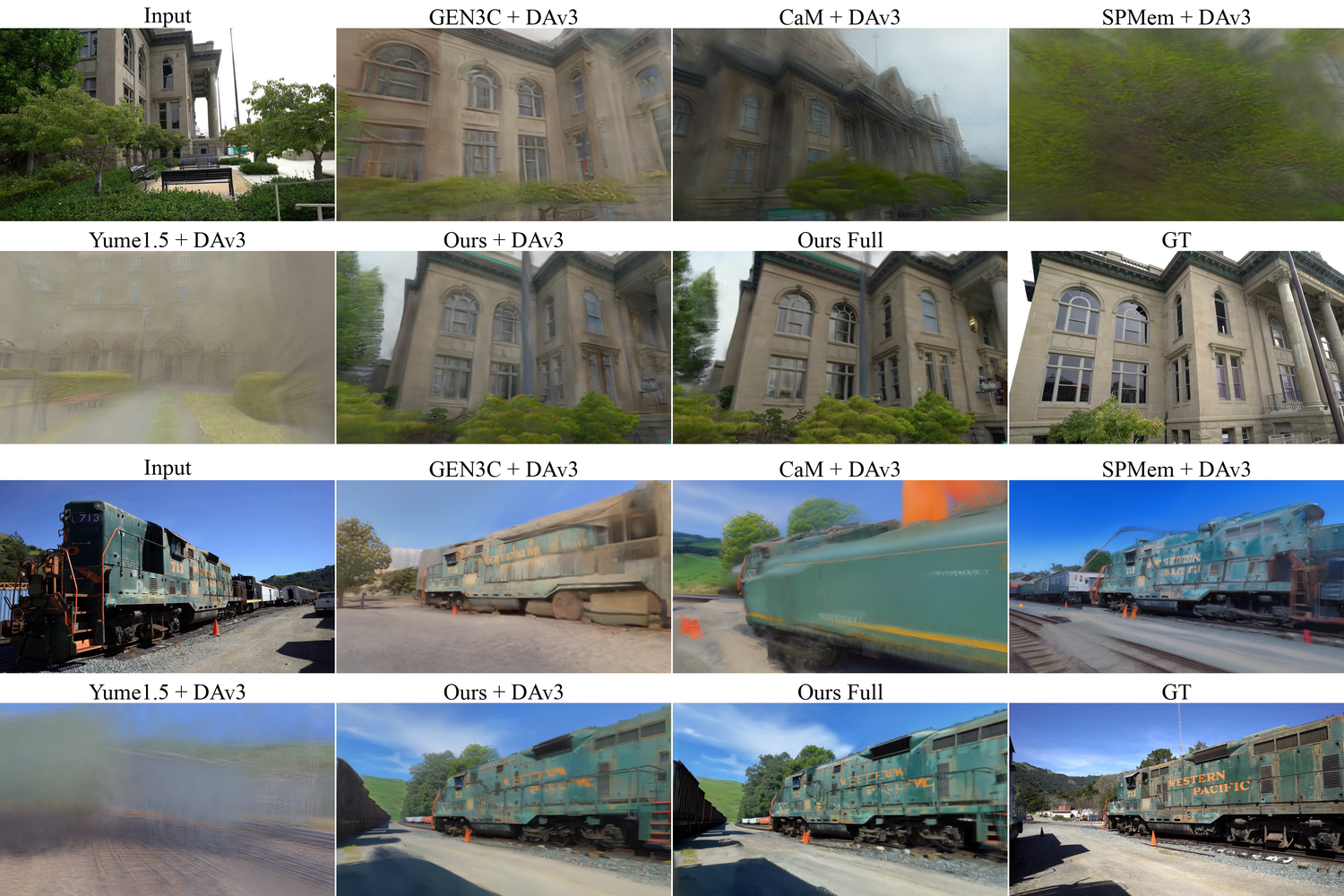}
  \caption{\textbf{3DGS comparisons.} We compare renderings from 3DGS scenes reconstructed from video diffusion model outputs, starting from a single input image from Tanks and Temples.}
  \label{fig:comparisons_3dgs}
\end{figure*}

\subsection{Evaluation on 3D Scene Generation}
\label{sec:exp_3d_scene}

\noindent\textbf{Baselines and Metrics.}
In this work, we focus on large-scale 3D scene generation. To construct competitive baselines, we pair the long video generation methods from Sec.~\ref{sec:exp_long_video} with Depth Anything V3~\cite{lin2025depth}, a state-of-the-art 3D reconstruction model that converts videos into 3DGS. We render novel views from the reconstructed 3DGS and evaluate with FID and Subjective Quality Score. We additionally report two LPIPS variants: LPIPS-G, computed between rendered novel views and ground-truth frames, which measures overall reconstruction quality; and LPIPS-P, computed between rendered novel views and the generated video frames, which quantifies the 3D consistency of the underlying video model---a more consistent video yields a more faithful 3D reconstruction and thus lower LPIPS-P. We also compare with prior generative reconstruction methods, Lyra~\cite{bahmani2025lyra} and FantasyWorld~\cite{dai2025fantasyworld}, which generate short videos and lift them to 3D but are inherently limited in scene scale.

\noindent\textbf{Quantitative Comparison.}
As shown in Tab.~\ref{tab:scene_gen}, our method achieves the best results across all metrics on both datasets. Both our variants---Ours + DAv3 and Ours Full---substantially outperform all baselines in LPIPS-G, FID, and Subjective Quality, demonstrating that the 3D consistency of our generated videos translates directly into higher-quality scene reconstructions. Furthermore, Ours Full consistently outperforms Ours + DAv3 across all metrics, validating the benefit of fine-tuning the reconstruction model on our generated scenes to improve robustness to generative artifacts. Notably, our method also achieves substantially lower LPIPS-P, confirming that our video model produces inherently more 3D-consistent outputs: the generated videos can be more faithfully reconstructed in 3D and re-rendered from novel viewpoints with minimal discrepancy.

\begin{figure*}[t]
  \centering
  \includegraphics[width=\linewidth]{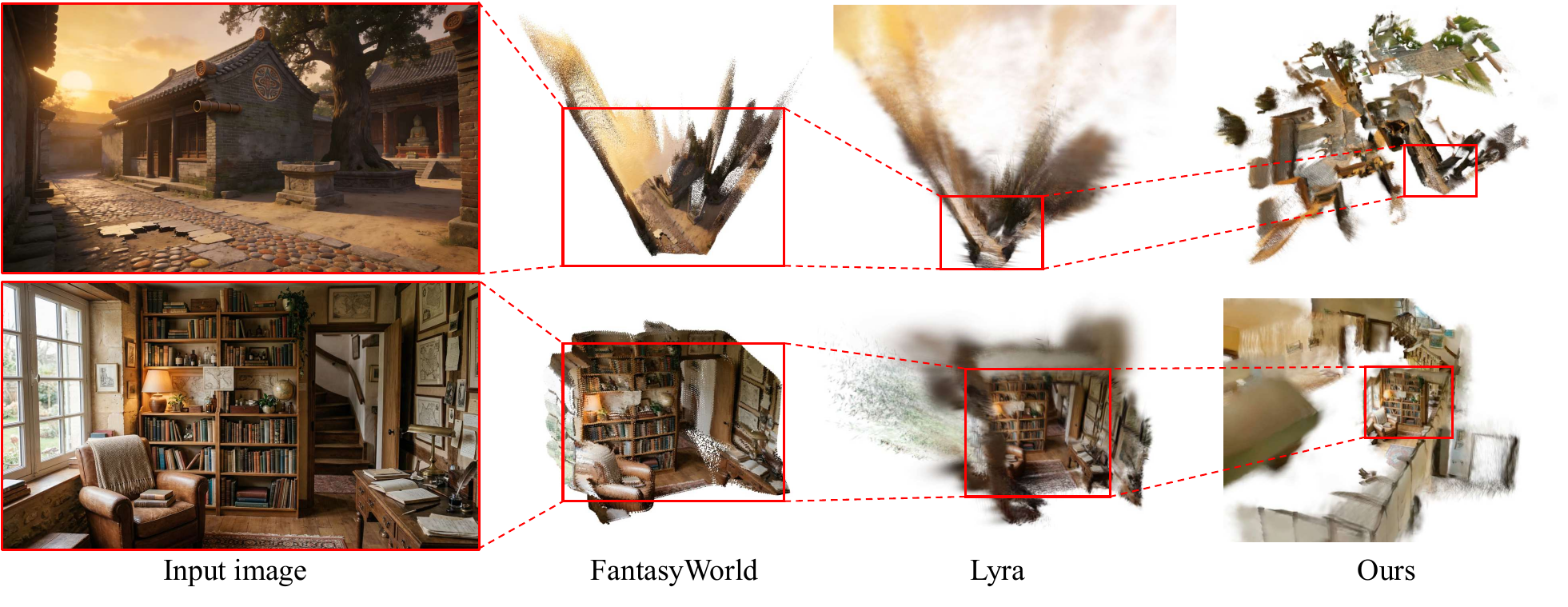}
  \vspace{-2em}
  \caption{\textbf{Qualitative comparison with Lyra and FantasyWorld.} We show 3DGS renderings (Lyra and Ours) and point cloud renderings (FantasyWorld) in bird's-eye view. Red bounding boxes highlight approximately the same spatial region across methods. Our interactive exploration framework produces scenes of significantly greater scale and complexity.}
  \label{fig:supp_lyra_fw}
\end{figure*}

\begin{table}[t]
  \caption{Ablation study on Tanks and Temples. We ablate key design choices of our framework.
    Best results are shown in \textbf{bold}.
  }
  \label{tab:ablation}
  \centering
  \resizebox{\linewidth}{!}{
  \begin{tabular}{@{}lccccccc@{}}
    \toprule
    Method & SSIM$\uparrow$ & LPIPS$\downarrow$ & FID$\downarrow$ & Subjective Qual.$\uparrow$ & Style Consist.$\uparrow$ & Camera Ctrl.$\uparrow$ & Reproj. Err.$\downarrow$ \\
    \midrule
    Ours & \textbf{0.384} & 0.552 & 51.33 & 43.35 & \textbf{85.07} & \textbf{63.87} & 0.069 \\
    w/ Global Point Cloud & 0.368 & 0.562 & 52.54 & 44.58 & 82.42 & 49.86 & 0.067 \\
    w/ Explicit Corr. Fusion & 0.370 & 0.554 & \textbf{49.13} & 45.71 & 83.28 & 57.29 & 0.071 \\
    w/o FramePack  & 0.362 & \textbf{0.549} & 50.98 & 45.27 & 80.61 & 62.62 & 0.079 \\
    w/o Self-Augmentation  & 0.363 & 0.568 & 55.15 & \textbf{47.88} & 77.98 & 53.92 & \textbf{0.066} \\
    \bottomrule
  \end{tabular}
  }
\end{table}

\begin{figure*}[t]
  \centering
  \includegraphics[width=\linewidth]{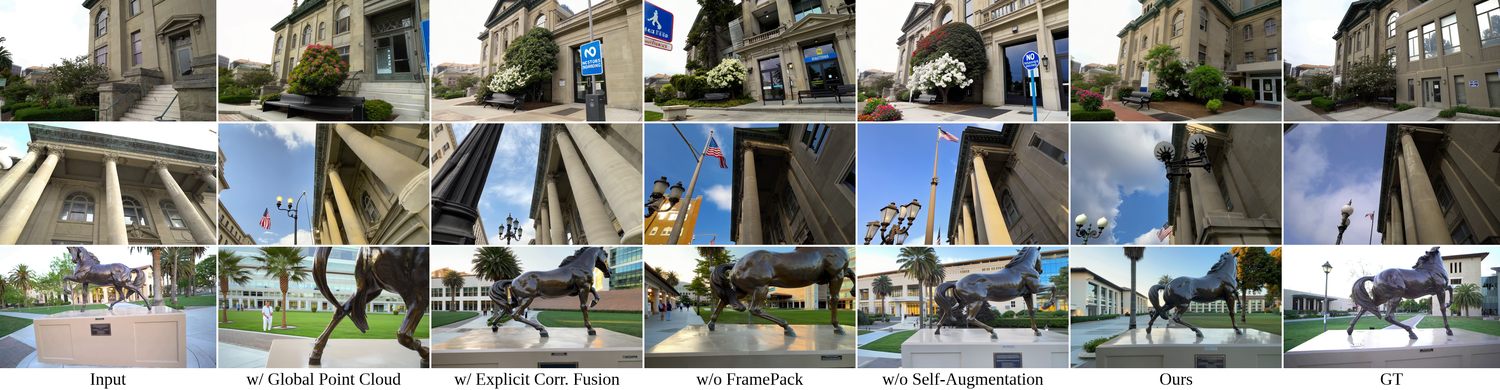}
  \caption{\textbf{Qualitative ablation study.} Given a single input image, we compare generations from our full model and ablated variants on Tanks and Temples scenes.}
  \label{fig:supp_ablation}
\end{figure*}

\noindent\textbf{Qualitative Comparison.}
We compare renderings of 3DGS scenes generated from single images in Fig.~\ref{fig:comparisons_3dgs}. While all baselines produce scenes with artifacts and floaters, our pipeline is able to generate realistic 3D scenes with high fidelity.
We further compare with Lyra~\cite{bahmani2025lyra} and FantasyWorld~\cite{dai2025fantasyworld} in Fig.~\ref{fig:supp_lyra_fw}. Both methods generate short videos and lift them to 3D, inherently limiting the achievable scene scale. In contrast, our interactive exploration framework allows users to iteratively define camera trajectories and progressively expand the environment, producing scenes of substantially greater spatial extent and complexity.

\subsection{Ablation Study}
\label{sec:ablation}

We ablate the key design choices of our framework on Tanks and Temples. Quantitative results are reported in Tab.~\ref{tab:ablation} and qualitative comparisons are shown in Fig.~\ref{fig:supp_ablation}.

\noindent\textbf{w/ Global Point Cloud} fuses all history frames into a single accumulated point cloud and conditions generation on its rendered images, replacing both the per-frame 3D cache and the correspondence-based conditioning. This significantly degrades Camera Controllability ($49.86$ vs.\ $63.87$) and Style Consistency ($82.42$ vs.\ $85.07$), confirming that accumulated depth errors corrupt the conditioning signal over long horizons. As shown in Fig.~\ref{fig:supp_ablation}, this variant produces noticeably inaccurate camera poses.

\noindent\textbf{w/ Explicit Corr.\ Fusion} replaces our learned MLP aggregation with explicit depth-reasoning-based fusion for merging correspondences from multiple source frames. Camera Controllability drops ($57.29$ vs.\ $63.87$), showing that learned aggregation handles noisy depth estimates more gracefully than hard geometric fusion.

\noindent\textbf{w/o FramePack} removes the FramePack temporal slots. Without temporal grounding, the model is prone to drifting, significantly reducing Style Consistency ($80.61$ vs.\ $85.07$) and increasing Reprojection Error ($0.079$ vs.\ $0.069$). As shown in Fig.~\ref{fig:supp_ablation}, this variant exhibits pronounced visual drifting.

\noindent\textbf{w/o Self-Augmentation} removes the self-augmentation training strategy. While per-frame Subjective Quality improves ($47.88$ vs.\ $43.35$), long-range consistency degrades substantially: Style Consistency drops to $77.98$ and Camera Controllability to $53.92$. Without exposure to imperfect conditioning during training, the model becomes brittle at inference when conditioning on its own imperfect outputs, causing errors to compound across segments, as visible in Fig.~\ref{fig:supp_ablation}.

\subsection{Applications}
\label{sec:applications}

Beyond quantitative evaluation, we demonstrate the practical applicability of our framework through an interactive GUI, in-the-wild scene generation, and downstream simulation.

\begin{figure*}[t]
  \centering
  \includegraphics[width=\linewidth]{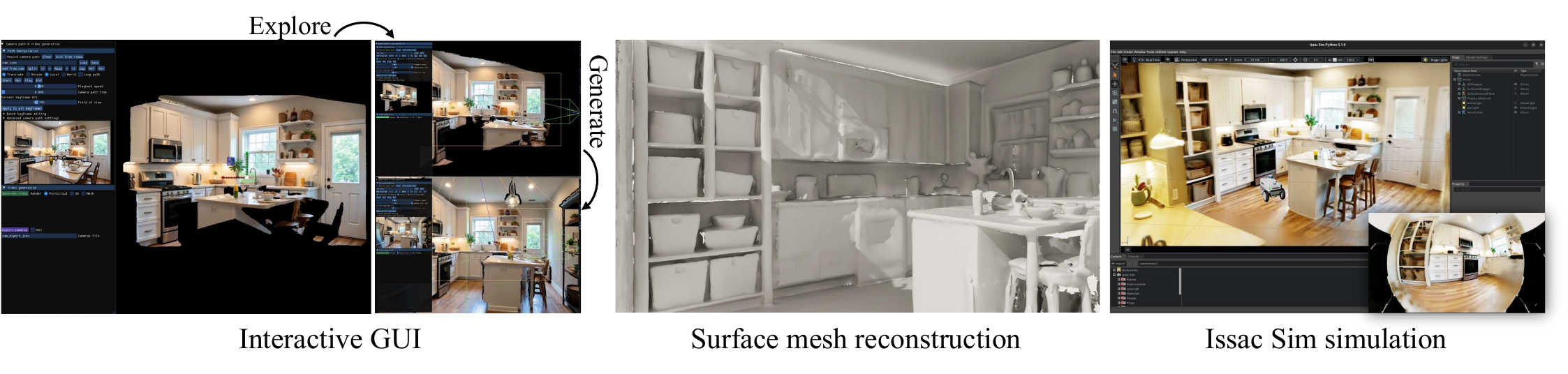}
  \vspace{-2em}
  \caption{\textbf{Applications.} Our interactive interface allows users to specify camera trajectories within the 3D cache to easily generate novel viewpoints. Moreover, the reconstructed 3DGS scenes can be converted into surface meshes and integrated into embodied AI simulators such as NVIDIA Isaac Sim for robot simulation.}
  \label{fig:applications}
\end{figure*}

\noindent\textbf{Interactive GUI.}
We build an interactive interface that allows users to specify camera trajectories within the 3D cache and progressively generate and explore scenes in real time, as shown in Fig.~\ref{fig:applications}. The GUI visualizes the accumulated point clouds, enabling users to plan trajectories that revisit previously explored regions or venture into unobserved areas. 

\begin{figure*}[t]
  \centering
  \includegraphics[width=\linewidth]{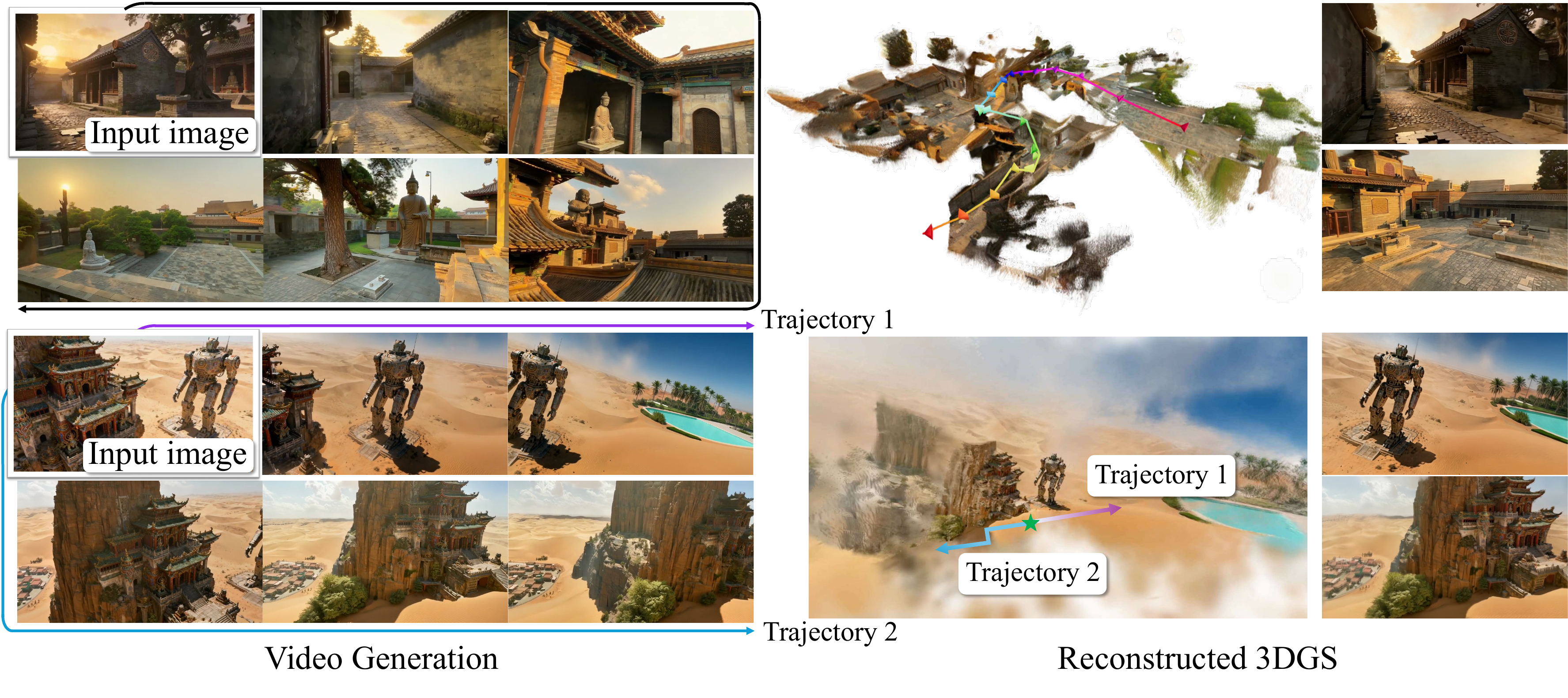}
  \vspace{-2em}
  \caption{\textbf{In-the-Wild Scene Generation.} We show video generations and 3DGS reconstructions for challenging in-the-wild input images that go beyond the training data distribution. Our approach supports flexible camera trajectories specified in the GUI for world exploration, including combining multiple trajectories from the same starting point (see second example).} %
  \label{fig:in-the-wild}
\end{figure*}

\noindent\textbf{In-the-Wild Scene Generation.}
We showcase our method on diverse in-the-wild images beyond the evaluation benchmarks, generating large-scale explorable 3D scenes from a single input image. As shown in Fig.~\ref{fig:teaser} and Fig.~\ref{fig:in-the-wild}, our framework produces globally consistent long videos and high-quality 3D reconstructions across a variety of scene types, including both indoor and outdoor environments.

\noindent\textbf{Embodied AI Simulation.}
The 3D Gaussian Splatting representations and meshes generated by our pipeline can be directly exported to physics engines for downstream applications. We demonstrate this by importing our reconstructed scenes into NVIDIA Isaac Sim, enabling physically grounded robot navigation and interaction within the generated environments. This highlights the potential of our framework for scalable embodied AI simulation without the need for real-world 3D data acquisition.

\section{Discussion}
\label{sec:conclusion}

In this work, we introduced \ours{}, a generative reconstruction framework that enables the creation of large-scale, explorable 3D environments.
Our approach addresses the key challenge of long-horizon consistency in camera-controlled video generation through dedicated anti-forgetting and anti-drifting mechanisms, and improves the reconstruction model to be robust to small generative errors. The generated scenes can be directly deployed for interactive exploration, virtual reality experiences, and simulation. 

Despite these advances, several limitations remain. First, our current framework focuses on static environments and does not explicitly model dynamic scenes, which remains an important direction for future work. Second, our video generation model inherits characteristics of the training data. In particular, the DL3DV dataset contains exposure variations across views, which the model may reproduce during generation. Such photometric inconsistencies can lead to artifacts in the feed-forward 3DGS reconstruction. Addressing photometric stability within the network~\cite{deutsch2026ppisp} or using photometrically consistent synthetic datasets~\cite{yu2025context}, e.g., from game engines, could lead to more consistent 3D scenes.

\section*{Acknowledgement}
We would like to thank Product Managers Aditya Mahajan and Matt Cragun for their valuable guidance and support. We also thank Oliver Hahn, David Pankratz, Christian Laforte, Gene Liu, and Rafal Karp for insightful discussions and feedback.
We are grateful to Yifeng Jiang, Nicolas Moenne-Loccoz, Tanki Zhang, Aditya Gupta, and Gavriel State for their prompt and helpful support in developing the Isaac Sim demo.
Finally, we sincerely acknowledge Merlin Nimier-David, Thomas Müller, and Alex Keller for their foundational interactive GUI, upon which our system builds

\clearpage
\newpage
\appendix

\section{Implementation Details}
\label{sec:supp_impl}

\subsection{Model Architecture}
\label{sec:supp_arch}

\noindent\textbf{Base Model.}
We build upon the Wan~2.1-14B DiT~\cite{wan2025wan} as our backbone video diffusion model. The VAE encodes videos at $8{\times}$ spatial and $4{\times}$ temporal downsampling with a latent channel dimension $C{=}16$. All training and inference are performed at a resolution of $832{\times}480$ pixels.

\noindent\textbf{Camera Conditioning Modules.}
We inject camera information through two complementary modules:
\begin{itemize}[leftmargin=*,nosep]
    \item \emph{Depth-warped conditioning}: We forward-warp the most recent frame to each target viewpoint using the estimated depth map, encode through the VAE, and concatenate with the denoising latent along the channel dimension.
    \item \emph{Pl\"ucker ray injection}: 6D Pl\"ucker ray coordinates are computed per pixel for all frames (temporal history, spatial memory, and generation tokens). These are projected to the DiT's hidden dimension via a pixel-shuffle layer followed by a single linear layer, yielding per-token ray embeddings $\mathbf{p}$. These are added to the token features before the query and key projections at every transformer block, \ie, $\mathbf{q} = W_Q(\mathbf{x} + \mathbf{p})$ and $\mathbf{k} = W_K(\mathbf{x} + \mathbf{p})$, while the value projection remains unmodified.
\end{itemize}

\noindent\textbf{Canonical Coordinate Injection.}
The forward-warped 4-channel canonical coordinate maps $[\hat{\mathbf{C}}_j;\, \hat{D}_j]$ are downsampled to match the latent spatial resolution via pixel shuffle. Each channel is encoded with sinusoidal positional encoding, and the resulting embeddings are aggregated through a pixel-shuffle layer followed by a single linear layer. The output is injected into the queries and keys of self-attention at every transformer block, but not the values, following the same injection scheme as the Pl\"ucker ray embeddings described above. This design ensures that the correspondence signal guides \emph{which} generation tokens attend to \emph{which} spatial slots, while the values remain unmodified from the pretrained model.

\begin{figure}[h]
    \centering
    \includegraphics[width=0.75\textwidth]{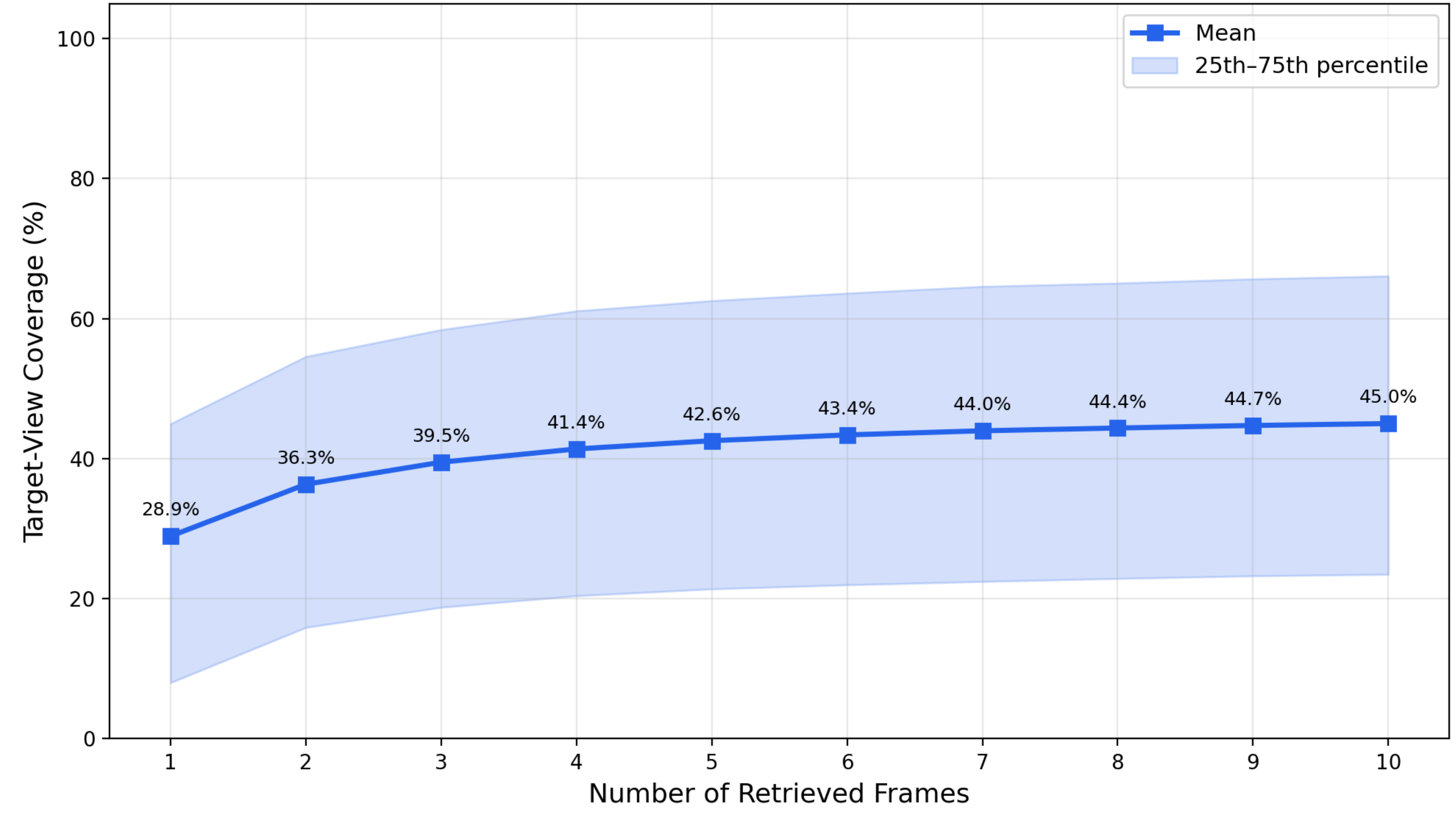}
    \caption{Target-frame coverage vs.\ number of retrieved spatial memory frames. We evaluate on training videos by treating the latter half as the target generation segment. Coverage is computed by forward-warping each retrieved frame to every target viewpoint using ground-truth depth; a target pixel is counted as covered only when the depth discrepancy between the warped point and the target ground-truth depth falls below a threshold. $N_s{=}5$ offers a favorable balance between spatial coverage and computational efficiency.}
    \label{fig:ns_ablation}
\end{figure}

\noindent\textbf{Number of Spatial Slots.}
We analyze the effect of the number of retrieved spatial memory frames $N_s$ on target-frame coverage in Fig.~\ref{fig:ns_ablation}. $N_s{=}5$ provides a good trade-off between coverage of previously visited regions and inference efficiency.

\subsection{Training}
\label{sec:supp_training}

\noindent\textbf{Spatial Memory.}
We retrieve $N_s{=}5$ spatial memory frames per autoregressive step. The downsampled point cloud in the 3D cache uses a subsampling factor of $d{=}8$. The visibility score occlusion threshold is $\delta{=}0.1$ (in normalized depth units).

\noindent\textbf{Self-Augmentation.}
We set the augmentation probability $p_{\text{aug}} = 0.7$.

\noindent\textbf{Optimization.}
We use AdamW~\cite{AdamW} with a learning rate of $3{\times}10^{-5}$ and weight decay $0.1$. Training uses a batch size of $64$ across $64$ NVIDIA GB200 GPUs. We train for $7{,}000$ iterations. All newly added modules are initialized with zero weights so that the model starts from the pretrained Wan~2.1 behavior. We use bf16 mixed-precision training throughout. 

\noindent\textbf{Flow Matching.}
We use rectified flow matching. During training we sample timesteps from a logit-normal distribution (mean 0, std 1 in logit space) with uniform time weighting; at inference we use the FlowUniPC~\cite{zhao2023unipc} multistep scheduler with 35 steps.

\subsection{Inference}
\label{sec:supp_inference}

\noindent\textbf{Classifier-Free Guidance.}
We apply classifier-free guidance (CFG) with a scale of $5.0$ for the text prompt.

\noindent\textbf{Runtime.}
Each autoregressive step (80 frames) takes approximately $194$ seconds on a single NVIDIA GB200 GPU for the full model (35 steps with CFG), including depth estimation, spatial memory retrieval, and DiT denoising. With Ours DMD (4 steps, no CFG), this reduces to approximately $15$ seconds per step. Spatial memory retrieval takes less than $1$ second per step in both cases.

\subsection{3D Reconstruction}
\label{sec:supp_recon}

\noindent\textbf{3DGS.}
The Gaussian DPT head downsampling factor is $k{=}2$, reducing the Gaussian count by $4{\times}$. To construct the fine-tuning dataset, we autoregressively generate 3{,}000 one-minute videos using images and camera trajectories from DL3DV~\cite{ling2024dl3dv}. We then fine-tune DAv3 on these scenes for $10{,}000$ iterations with a learning rate of $5{\times}10^{-5}$ and batch size $8$.

\noindent\textbf{Mesh Extraction.}
The mesh extraction step extracts a triangular mesh of the scene using a hierarchical sparse grid. The number of levels and voxel sizes for each level in the hierarchy are determined by the scale of the scene. The depth from the Gaussian reconstruction is used to compute a signed distance field, and a single surface mesh is extracted by running marching cubes on each level and merging them at level transitions.

\subsection{Related Work}
\label{sec:supp_related}

We provide more extensive related work discussion in addition to Sec.~\ref{sec:related}.

\noindent\textbf{3D generation.}
Early work on 3D generation largely focused on category-specific object synthesis, extending GAN-based frameworks to 3D by incorporating neural rendering as an inductive bias~\citep{devries2021unconstrained, chan2022efficient, or2022stylesdf, schwarz2022voxgraf, bahmani2023cc3d, gao2022get3d}. The introduction of CLIP-based supervision~\citep{radford2021learning} enabled more flexible generation pipelines, supporting both text-conditioned synthesis and semantic editing~\citep{chen2018text2shape, jain2022zero, sanghi2022clip, jetchev2021clipmatrix, gao2023textdeformer, wang2022clip}. More recently, diffusion-based methods have substantially improved visual fidelity by replacing CLIP guidance with Score Distillation Sampling (SDS)~\citep{poole2022dreamfusion, wang2023prolificdreamer, lin2022magic3d, chen2023fantasia3d, liang2023luciddreamer, wang2023score, li2024controllable, he2024gvgen, ye2024dreamreward, liu2023humangaussian, yu2023text, katzir2023noise, lee2023dreamflow, sun2023dreamcraft3d}.
To improve geometric consistency, a number of approaches explicitly enforce multi-view coherence by generating or supervising across multiple viewpoints~\citep{lin2023consistent123, liu2023zero, shi2023mvdream, feng2024fdgaussian, liu2024isotropic3d, kim2023neuralfield, voleti2024sv3d, hollein2024viewdiff, tang2024pixel, gao2024cat3d,wang2025act,kant2025pippo,yuan2025generative,ren2024xcube}. In parallel, some methods formulate scene generation as an iterative inpainting process to progressively expand 3D environments~\citep{hollein2023text2room, shriram2024realmdreamer}. Another line of work lifts 2D observations into 3D representations using NeRF~\citep{NeRF}, 3D Gaussian Splatting~\citep{kerbl20233d}, or mesh-based formulations in combination with diffusion priors~\citep{chan2023generative, tang2023make, gu2023nerfdiff, liu2023syncdreamer, yoo2023dreamsparse, tewari2024diffusion, qian2023magic123, long2023wonder3d, wan2023cad, szymanowicz2023viewset,lu2024infinicube}.

\noindent\textbf{Feed-forward 3D models.}
A complementary line of research focuses on feed-forward architectures that directly infer 3D structure from images or text in a single pass~\citep{hong2023lrm, li2023instant3d, xu2023dmv3d, xu2024grm, zhang2024compress3d, han2024vfusion3d, jiang2024brightdreamer, xie2024latte3d, tang2024lgm, tochilkin2024triposr, qian2024atom, szymanowicz2023splatter, szymanowicz2024flash3d, liang2024wonderland, szymanowicz2025bolt3d,schwarz2025generative,yang2025matrix,zhang2025spatialcrafter}. While these methods enable efficient 3D generation, they are generally restricted to static scene representations. Other approaches specialize in narrow domains such as facial reconstruction~\citep{kirschstein2025avat3r}. Some works~\citep{liang2024btimer,xu20254dgt} address real-world dynamic scenes, but struggle to generalize to diverse generated content or large viewpoint variations.

\clearpage
{
\small
\bibliographystyle{abbrv}
\bibliography{main}

@String(CVPR  = {IEEE Conf. Comput. Vis. Pattern Recog.})

@String(ICCV  = {Int. Conf. Comput. Vis.})

@String(ECCV  = {Eur. Conf. Comput. Vis.})

@String(NeurIPS = {Adv. Neural Inform. Process. Syst.})

@String(ICML  = {Int. Conf. Mach. Learn.})

@String(ICLR  = {Int. Conf. Learn. Represent.})

@String(ACCV  = {Asian Conf. Comput. Vis.})

@String(TOG   = {ACM Trans. Graph.})

@String(CVPR  = {CVPR})

@String(ICCV  = {ICCV})

@String(ECCV  = {ECCV})

@String(NeurIPS = {NeurIPS})

@String(ICML  = {ICML})

@String(ICLR  = {ICLR})

@String(ACCV  = {ACCV})

@String(TOG   = {ACM TOG})

@inproceedings{ren2025gen3c,
  title={Gen3c: 3d-informed world-consistent video generation with precise camera control},
  author={Ren, Xuanchi and Shen, Tianchang and Huang, Jiahui and Ling, Huan and Lu, Yifan and Nimier-David, Merlin and M{\"u}ller, Thomas and Keller, Alexander and Fidler, Sanja and Gao, Jun},
  booktitle=CVPR,
  pages={6121--6132},
  year={2025}
}

@article{mao2025yume,
  title={Yume-1.5: A Text-Controlled Interactive World Generation Model},
  author={Mao, Xiaofeng and Li, Zhen and Li, Chuanhao and Xu, Xiaojie and Ying, Kaining and He, Tong and Pang, Jiangmiao and Qiao, Yu and Zhang, Kaipeng},
  journal={arXiv preprint arXiv:2512.22096},
  year={2025}
}

@inproceedings{yu2025context,
  title={Context as memory: Scene-consistent interactive long video generation with memory retrieval},
  author={Yu, Jiwen and Bai, Jianhong and Qin, Yiran and Liu, Quande and Wang, Xintao and Wan, Pengfei and Zhang, Di and Liu, Xihui},
  booktitle={SIGGRAPH Asia},
  pages={1--11},
  year={2025}
}

@article{wu2025video,
  title={Video world models with long-term spatial memory},
  author={Wu, Tong and Yang, Shuai and Po, Ryan and Xu, Yinghao and Liu, Ziwei and Lin, Dahua and Wetzstein, Gordon},
  journal={arXiv preprint arXiv:2506.05284},
  year={2025}
}

@inproceedings{li2025vmem,
  title={Vmem: Consistent interactive video scene generation with surfel-indexed view memory},
  author={Li, Runjia and Torr, Philip and Vedaldi, Andrea and Jakab, Tomas},
  booktitle=ICCV,
  pages={25690--25699},
  year={2025}
}

@article{huang2025vipe,
  title={Vipe: Video pose engine for 3d geometric perception},
  author={Huang, Jiahui and Zhou, Qunjie and Rabeti, Hesam and Korovko, Aleksandr and Ling, Huan and Ren, Xuanchi and Shen, Tianchang and Gao, Jun and Slepichev, Dmitry and Lin, Chen-Hsuan and others},
  journal={arXiv preprint arXiv:2508.10934},
  year={2025}
}

@inproceedings{ling2024dl3dv,
  title={Dl3dv-10k: A large-scale scene dataset for deep learning-based 3d vision},
  author={Ling, Lu and Sheng, Yichen and Tu, Zhi and Zhao, Wentian and Xin, Cheng and Wan, Kun and Yu, Lantao and Guo, Qianyu and Yu, Zixun and Lu, Yawen and others},
  booktitle=CVPR,
  pages={22160--22169},
  year={2024}
}

@article{lin2025depth,
  title={Depth anything 3: Recovering the visual space from any views},
  author={Lin, Haotong and Chen, Sili and Liew, Junhao and Chen, Donny Y and Li, Zhenyu and Shi, Guang and Feng, Jiashi and Kang, Bingyi},
  journal={arXiv preprint arXiv:2511.10647},
  year={2025}
}

@inproceedings{bahmani2025lyra,
  title={Lyra: Generative 3D Scene Reconstruction via Video Diffusion Model Self-Distillation},
  author={Bahmani, Sherwin and Shen, Tianchang and Ren, Jiawei and Huang, Jiahui and Jiang, Yifeng and Turki, Haithem and Tagliasacchi, Andrea and Lindell, David B. and Gojcic, Zan and Fidler, Sanja and Ling, Huan and Gao, Jun and Ren, Xuanchi},
  booktitle=ICLR,
  year={2026}
}

@inproceedings{dai2025fantasyworld,
  title={FantasyWorld: Geometry-Consistent World Modeling via Unified Video and 3D Prediction},
  author={Dai, Yixiang and Jiang, Fan and Wang, Chiyu and Xu, Mu and Qi, Yonggang},
  booktitle=ICLR,
  year={2026}
}

@article{knapitsch2017tanks,
  title={Tanks and Temples: Benchmarking Large-Scale Scene Reconstruction},
  author={Knapitsch, Arno and Park, Jaesik and Zhou, Qian-Yi and Koltun, Vladlen},
  journal=TOG,
  volume={36},
  number={4},
  year={2017}
}

@inproceedings{duan2025worldscore,
  title={WorldScore: A Unified Evaluation Benchmark for World Generation},
  author={Duan, Haoyi and Yu, Hong-Xing and Chen, Sirui and Fei-Fei, Li and Wu, Jiajun},
  booktitle=ICCV,
  year={2025}
}

@article{wan2025wan,
  title={Wan: Open and Advanced Large-Scale Video Generative Models},
  author={Team Wan and Wang, Ang and Ai, Baole and Wen, Bin and Mao, Chaojie and Xie, Chen-Wei and Chen, Di and Yu, Feiwu and Zhao, Haiming and Yang, Jianxiao and others},
  journal={arXiv preprint arXiv:2503.20314},
  year={2025}
}

@article{bai2025qwen3vl,
  title={Qwen3-VL Technical Report},
  author={Qwen Team},
  journal={arXiv preprint arXiv:2511.21631},
  year={2025}
}

@article{ball2025genie,
  title={Genie 3: A new frontier for world models},
  author={Ball, Philip J and Bauer, Jakob and Belletti, Frank and Brownfield, B and Ephrat, A and Fruchter, S and Gupta, A and Holsheimer, K and Holynski, A and Hron, J and others},
  journal={Google DeepMind Blog},
  pages={253--279},
  year={2025}
}

@inproceedings{chen2023ray,
  title={Ray conditioning: Trading photo-consistency for photo-realism in multi-view image generation},
  author={Chen, Eric Ming and Holalkere, Sidhanth and Yan, Ruyu and Zhang, Kai and Davis, Abe},
  booktitle=ICCV,
  year={2023}
}

@inproceedings{FID,
  title={Gans trained by a two time-scale update rule converge to a local nash equilibrium},
  author={Heusel, Martin and Ramsauer, Hubert and Unterthiner, Thomas and Nessler, Bernhard and Hochreiter, Sepp},
  booktitle={Proc. NeurIPS},
  year={2017}
}

@inproceedings{NeRF,
  title={Nerf: Representing scenes as neural radiance fields for view synthesis},
  author={Mildenhall, Ben and Srinivasan, Pratul P and Tancik, Matthew and Barron, Jonathan T and Ramamoorthi, Ravi and Ng, Ren},
  booktitle={Proc. ECCV},
  year={2020},
}

@inproceedings{LPIPS,
  title={The Unreasonable Effectiveness of Deep Features as a Perceptual Metric},
  author={Zhang, Richard and Isola, Phillip and Efros, Alexei A and Shechtman, Eli and Wang, Oliver},
  booktitle={CVPR},
  year={2018}
}

@inproceedings{radford2021learning,
  title={Learning transferable visual models from natural language supervision},
  author={Radford, Alec and Kim, Jong Wook and Hallacy, Chris and Ramesh, Aditya and Goh, Gabriel and Agarwal, Sandhini and Sastry, Girish and Askell, Amanda and Mishkin, Pamela and Clark, Jack and others},
  booktitle={Proc. ICML},
  year={2021},
}

@inproceedings{MotionCtrl,
  title={MotionCtrl: A Unified and Flexible Motion Controller for Video Generation},
  author={Wang, Zhouxia and Yuan, Ziyang and Wang, Xintao and Chen, Tianshui and Xia, Menghan and Luo, Ping and Shan, Yin},
  booktitle={SIGGRAPH},
  year={2024}
}

@article{cameractrl,
    title={CameraCtrl: Enabling Camera Control for Text-to-Video Generation},
    author={Hao He and Yinghao Xu and Yuwei Guo and Gordon Wetzstein and Bo Dai and Hongsheng Li and Ceyuan Yang},
    year={2024},
    journal={arXiv preprint arXiv:2404.02101},
}

@article{Sora,
  title={Video generation models as world simulators},
  author={Tim Brooks and Bill Peebles and Connor Holmes and Will DePue and Yufei Guo and Li Jing and David Schnurr and Joe Taylor and Troy Luhman and Eric Luhman and Clarence Ng and Ricky Wang and Aditya Ramesh},
  year={2024},
  url={https://openai.com/research/video-generation-models-as-world-simulators},
  journal={OpenAI technical reports}
}

@misc{veo,
  title={Veo},
author = {Abhishek Sharma and Adams Yu and Ali Razavi and Andeep Toor and Andrew Pierson and Ankush Gupta and Austin Waters and Daniel Tanis and Dumitru Erhan and Eric Lau and Eleni Shaw and Gabe Barth-Maron and Greg Shaw and Han Zhang and Henna Nandwani and Hernan Moraldo and Hyunjik Kim and Irina Blok and Jakob Bauer and Jeff Donahue and Junyoung Chung and Kory Mathewson and Kurtis David and Lasse Espeholt and Marc van Zee and Matt McGill and Medhini Narasimhan and Miaosen Wang and Mikołaj Bińkowski and Mohammad Babaeizadeh and Mohammad Taghi Saffar and Nick Pezzotti and Pieter-Jan Kindermans and Poorva Rane and Rachel Hornung and Robert Riachi and Ruben Villegas and Rui Qian and Sander Dieleman and Serena Zhang and Serkan Cabi and Shixin Luo and Shlomi Fruchter and Signe Nørly and Srivatsan Srinivasan and Tobias Pfaff and Tom Hume and Vikas Verma and Weizhe Hua and William Zhu and Xinchen Yan and Xinyu Wang and Yelin Kim and Yuqing Du and Yutian Chen},
  year={2024},
  url={https://deepmind.google/technologies/veo/},
}

@inproceedings{kerbl20233d,
  title={3d gaussian splatting for real-time radiance field rendering},
  author={Kerbl, Bernhard and Kopanas, Georgios and Leimk{\"u}hler, Thomas and Drettakis, George},
  booktitle={ACM TOG},
  year={2023},
}

@article{xu2024camco,
  title={CamCo: Camera-Controllable 3D-Consistent Image-to-Video Generation},
  author={Xu, Dejia and Nie, Weili and Liu, Chao and Liu, Sifei and Kautz, Jan and Wang, Zhangyang and Vahdat, Arash},
  journal={arXiv preprint arXiv:2406.02509},
  year={2024}
}

@inproceedings{chan2023generative,
  title={Generative novel view synthesis with 3d-aware diffusion models},
  author={Chan, Eric R and Nagano, Koki and Chan, Matthew A and Bergman, Alexander W and Park, Jeong Joon and Levy, Axel and Aittala, Miika and De Mello, Shalini and Karras, Tero and Wetzstein, Gordon},
  booktitle={Proc. ICCV},
  year={2023}
}

@inproceedings{gao2024cat3d,
  title={Cat3d: Create anything in 3d with multi-view diffusion models},
  author={Gao, Ruiqi and Holynski, Aleksander and Henzler, Philipp and Brussee, Arthur and Martin-Brualla, Ricardo and Srinivasan, Pratul and Barron, Jonathan T and Poole, Ben},
  booktitle={Proc. NeurIPS},
  year={2024}
}

@inproceedings{sitzmann2021light,
  title={Light field networks: Neural scene representations with single-evaluation rendering},
  author={Sitzmann, Vincent and Rezchikov, Semon and Freeman, Bill and Tenenbaum, Josh and Durand, Fredo},
  booktitle={Proc. NeurIPS},
  year={2021}
}

@article{szymanowicz2024flash3d,
  title={Flash3D: Feed-Forward Generalisable 3D Scene Reconstruction from a Single Image},
  author={Szymanowicz, Stanislaw and Insafutdinov, Eldar and Zheng, Chuanxia and Campbell, Dylan and Henriques, Jo{\~a}o F and Rupprecht, Christian and Vedaldi, Andrea},
  journal={Proc. 3DV},
  year={2025}
}

@inproceedings{devries2021unconstrained,
  title={Unconstrained scene generation with locally conditioned radiance fields},
  author={DeVries, Terrance and Bautista, Miguel Angel and Srivastava, Nitish and Taylor, Graham W and Susskind, Joshua M},
  booktitle={Proc. ICCV},
  year={2021}
}

@inproceedings{or2022stylesdf,
  title={{StyleSDF}: High-resolution {3D}-consistent image and geometry generation},
  author={Or-El, Roy and Luo, Xuan and Shan, Mengyi and Shechtman, Eli and Park, Jeong Joon and Kemelmacher-Shlizerman, Ira},
  booktitle={Proc. CVPR},
  year={2022}
}

@inproceedings{chan2022efficient,
  title={Efficient geometry-aware {3D} generative adversarial networks},
  author={Chan, Eric R and Lin, Connor Z and Chan, Matthew A and Nagano, Koki and Pan, Boxiao and De Mello, Shalini and Gallo, Orazio and Guibas, Leonidas J and Tremblay, Jonathan and Khamis, Sameh and others},
  booktitle={Proc. CVPR},
  year={2022}
}

@inproceedings{schwarz2022voxgraf,
  title={{VoxGRAF}: Fast {3D}-aware image synthesis with sparse voxel grids},
  author={Schwarz, Katja and Sauer, Axel and Niemeyer, Michael and Liao, Yiyi and Geiger, Andreas},
  booktitle={Proc. NeurIPS},
  year={2022}
}

@inproceedings{bahmani2023cc3d,
  title={{CC3D}: Layout-conditioned generation of compositional {3D} scenes},
  author={Bahmani, Sherwin and Park, Jeong Joon and Paschalidou, Despoina and Yan, Xingguang and Wetzstein, Gordon and Guibas, Leonidas and Tagliasacchi, Andrea},
  booktitle={Proc. ICCV},
  year={2023}
}

@inproceedings{liu2023humangaussian,
  title={{HumanGaussian}: Text-driven {3D} human generation with {Gaussian} splatting},
  author={Liu, Xian and Zhan, Xiaohang and Tang, Jiaxiang and Shan, Ying and Zeng, Gang and Lin, Dahua and Liu, Xihui and Liu, Ziwei},
  booktitle={Proc. CVPR},
  year={2024}
}

@article{yu2023text,
  title={Text-to-{3D} with classifier score distillation},
  author={Yu, Xin and Guo, Yuan-Chen and Li, Yangguang and Liang, Ding and Zhang, Song-Hai and Qi, Xiaojuan},
  journal={arXiv preprint arXiv:2310.19415},
  year={2023}
}

@inproceedings{katzir2023noise,
  title={Noise-free score distillation},
  author={Katzir, Oren and Patashnik, Or and Cohen-Or, Daniel and Lischinski, Dani},
  booktitle={Proc. ICLR},
  year={2024}
}

@inproceedings{chen2018text2shape,
  title={{Text2Shape}: Generating shapes from natural language by learning joint embeddings},
  author={Chen, Kevin and Choy, Christopher B and Savva, Manolis and Chang, Angel X and Funkhouser, Thomas and Savarese, Silvio},
  booktitle={Proc. ACCV},
  year={2018},
}

@article{jetchev2021clipmatrix,
  title={{ClipMatrix}: Text-controlled creation of {3D} textured meshes},
  author={Jetchev, Nikolay},
  journal={arXiv preprint arXiv:2109.12922},
  year={2021}
}

@inproceedings{sanghi2022clip,
  title={{CLIP-Forge}: Towards zero-shot text-to-shape generation},
  author={Sanghi, Aditya and Chu, Hang and Lambourne, Joseph G and Wang, Ye and Cheng, Chin-Yi and Fumero, Marco and Malekshan, Kamal Rahimi},
  booktitle={Proc. CVPR},
  year={2022}
}

@inproceedings{jain2022zero,
  title={Zero-shot text-guided object generation with dream fields},
  author={Jain, Ajay and Mildenhall, Ben and Barron, Jonathan T and Abbeel, Pieter and Poole, Ben},
  booktitle={Proc. CVPR},
  year={2022}
}

@inproceedings{wang2022clip,
  title={Clip-{N}e{RF}: Text-and-image driven manipulation of neural radiance fields},
  author={Wang, Can and Chai, Menglei and He, Mingming and Chen, Dongdong and Liao, Jing},
  booktitle={Proc. CVPR},
  year={2022}
}

@inproceedings{gao2023textdeformer,
  title={{TextDeformer}: Geometry manipulation using text guidance},
  author={Gao, William and Aigerman, Noam and Groueix, Thibault and Kim, Vova and Hanocka, Rana},
  booktitle={SIGGRAPH},
  year={2023}
}

@article{tang2023make,
  title={Make-it-{3D}: High-fidelity {3D} creation from a single image with diffusion prior},
  author={Tang, Junshu and Wang, Tengfei and Zhang, Bo and Zhang, Ting and Yi, Ran and Ma, Lizhuang and Chen, Dong},
  journal={arXiv preprint arXiv:2303.14184},
  year={2023}
}

@inproceedings{liu2023zero,
  title={Zero-1-to-3: Zero-shot one image to {3D} object},
  author={Liu, Ruoshi and Wu, Rundi and Van Hoorick, Basile and Tokmakov, Pavel and Zakharov, Sergey and Vondrick, Carl},
  booktitle={Proc. ICCV},
  year={2023}
}

@inproceedings{gu2023nerfdiff,
  title={{NerfDiff}: Single-image view synthesis with {NeRF}-guided distillation from {3D}-aware diffusion},
  author={Gu, Jiatao and Trevithick, Alex and Lin, Kai-En and Susskind, Joshua M and Theobalt, Christian and Liu, Lingjie and Ramamoorthi, Ravi},
  booktitle={Proc. ICML},
  year={2023},
}

@inproceedings{liu2023syncdreamer,
  title={{SyncDreamer}: Generating Multiview-consistent Images from a Single-view Image},
  author={Liu, Yuan and Lin, Cheng and Zeng, Zijiao and Long, Xiaoxiao and Liu, Lingjie and Komura, Taku and Wang, Wenping},
  booktitle={Proc. ICLR},
  year={2024}
}

@inproceedings{yoo2023dreamsparse,
  title={{DreamSparse}: Escaping from {Plato's} Cave with {2D} Diffusion Model Given Sparse Views},
  author={Yoo, Paul and Guo, Jiaxian and Matsuo, Yutaka and Gu, Shixiang Shane},
  booktitle={arXiv preprint arXiv:2306.03414},
  year={2023}
}

@inproceedings{tewari2024diffusion,
  title={Diffusion with forward models: Solving stochastic inverse problems without direct supervision},
  author={Tewari, Ayush and Yin, Tianwei and Cazenavette, George and Rezchikov, Semon and Tenenbaum, Josh and Durand, Fr{\'e}do and Freeman, Bill and Sitzmann, Vincent},
  booktitle={Proc. NeurIPS},
  year={2023}
}

@inproceedings{lin2023consistent123,
  title={Consistent123: One Image to Highly Consistent {3D} Asset Using Case-Aware Diffusion Priors},
  author={Lin, Yukang and Han, Haonan and Gong, Chaoqun and Xu, Zunnan and Zhang, Yachao and Li, Xiu},
  booktitle={arXiv preprint arXiv:2309.17261},
  year={2023}
}

@inproceedings{li2023instant3d,
  title={{Instant3D}: Fast Text-to-{3D} with Sparse-View Generation and Large Reconstruction Model},
  author={Jiahao Li and Hao Tan and Kai Zhang and Zexiang Xu and Fujun Luan and Yinghao Xu and Yicong Hong and Kalyan Sunkavalli and Greg Shakhnarovich and Sai Bi},
  booktitle={Proc. ICLR},
  year={2024},
}

@inproceedings{hong2023lrm,
  title={{LRM}: Large Reconstruction Model for Single Image to {3D}},
  author={Hong, Yicong and Zhang, Kai and Gu, Jiuxiang and Bi, Sai and Zhou, Yang and Liu, Difan and Liu, Feng and Sunkavalli, Kalyan and Bui, Trung and Tan, Hao},
  booktitle={Proc. ICLR},
  year={2024}
}

@inproceedings{qian2023magic123,
  title={Magic123: One image to high-quality {3D} object generation using both {2D} and {3D} diffusion priors},
  author={Qian, Guocheng and Mai, Jinjie and Hamdi, Abdullah and Ren, Jian and Siarohin, Aliaksandr and Li, Bing and Lee, Hsin-Ying and Skorokhodov, Ivan and Wonka, Peter and Tulyakov, Sergey and others},
  booktitle={Proc. ICLR},
  year={2024}
}

@inproceedings{poole2022dreamfusion,
  author = {Poole, Ben and Jain, Ajay and Barron, Jonathan T. and Mildenhall, Ben},
  title = {Dream{F}usion: Text-to-{3D} using {2D} Diffusion},
  booktitle = {Proc. ICLR},
  year = {2023},
}

@inproceedings{lin2022magic3d,
  title={Magic3{D}: High-Resolution Text-to-{3D} Content Creation},
  author={Lin, Chen-Hsuan and Gao, Jun and Tang, Luming and Takikawa, Towaki and Zeng, Xiaohui and Huang, Xun and Kreis, Karsten and Fidler, Sanja and Liu, Ming-Yu and Lin, Tsung-Yi},
  booktitle={Proc. CVPR},
  year={2023}
}

@inproceedings{wang2023prolificdreamer,
  title={{ProlificDreamer}: High-Fidelity and Diverse Text-to-{3D} Generation with Variational Score Distillation},
  author={Wang, Zhengyi and Lu, Cheng and Wang, Yikai and Bao, Fan and Li, Chongxuan and Su, Hang and Zhu, Jun},
  booktitle={Proc. NeurIPS},
  year={2023}
}

@article{chen2023fantasia3d,
  title={{Fantasia3D}: Disentangling geometry and appearance for high-quality text-to-{3D} content creation},
  author={Chen, Rui and Chen, Yongwei and Jiao, Ningxin and Jia, Kui},
  journal={arXiv preprint arXiv:2303.13873},
  year={2023}
}

@inproceedings{shi2023mvdream,
  title={{MVDream}: Multi-view Diffusion for {3D} Generation},
  author={Shi, Yichun and Wang, Peng and Ye, Jianglong and Mai, Long and Li, Kejie and Yang, Xiao},
  booktitle={Proc. ICLR},
  year={2024}
}

@article{liang2023luciddreamer,
  title={Luciddreamer: Towards high-fidelity text-to-{3D} generation via interval score matching},
  author={Liang, Yixun and Yang, Xin and Lin, Jiantao and Li, Haodong and Xu, Xiaogang and Chen, Yingcong},
  journal={arXiv preprint arXiv:2311.11284},
  year={2023}
}

@inproceedings{xu2023dmv3d,
  title={{DMV3D}: Denoising multi-view diffusion using {3D} large reconstruction model},
  author={Xu, Yinghao and Tan, Hao and Luan, Fujun and Bi, Sai and Wang, Peng and Li, Jiahao and Shi, Zifan and Sunkavalli, Kalyan and Wetzstein, Gordon and Xu, Zexiang and others},
  booktitle={Proc. ICLR},
  year={2024}
}

@inproceedings{long2023wonder3d,
  title={{Wonder3D}: Single image to {3D} using cross-domain diffusion},
  author={Long, Xiaoxiao and Guo, Yuan-Chen and Lin, Cheng and Liu, Yuan and Dou, Zhiyang and Liu, Lingjie and Ma, Yuexin and Zhang, Song-Hai and Habermann, Marc and Theobalt, Christian and others},
  booktitle={Proc. CVPR},
  year={2024}
}

@inproceedings{wan2023cad,
  title={{CAD}: Photorealistic {3D} generation via adversarial distillation},
  author={Wan, Ziyu and Paschalidou, Despoina and Huang, Ian and Liu, Hongyu and Shen, Bokui and Xiang, Xiaoyu and Liao, Jing and Guibas, Leonidas},
  booktitle={Proc. CVPR},
  year={2024}
}

@inproceedings{lee2023dreamflow,
  title={{DreamFlow}: High-quality text-to-{3D} generation by Approximating Probability Flow},
  author={Lee, Kyungmin and Sohn, Kihyuk and Shin, Jinwoo},
  booktitle={Proc. ICLR},
  year={2024}
}

@article{qian2024atom,
  title={AToM: Amortized Text-to-Mesh using 2D Diffusion},
  author={Qian, Guocheng and Cao, Junli and Siarohin, Aliaksandr and Kant, Yash and Wang, Chaoyang and Vasilkovsky, Michael and Lee, Hsin-Ying and Fang, Yuwei and Skorokhodov, Ivan and Zhuang, Peiye and others},
  journal={arXiv preprint arXiv:2402.00867},
  year={2024}
}

@article{voleti2024sv3d,
  title={{SV3D}: Novel Multi-view Synthesis and {3D} Generation from a Single Image using Latent Video Diffusion},
  author={Voleti, Vikram and Yao, Chun-Han and Boss, Mark and Letts, Adam and Pankratz, David and Tochilkin, Dmitry and Laforte, Christian and Rombach, Robin and Jampani, Varun},
  journal={arXiv preprint arXiv:2403.12008},
  year={2024}
}

@article{tochilkin2024triposr,
  title={T{riposr}: Fast {3D} object reconstruction from a single image},
  author={Tochilkin, Dmitry and Pankratz, David and Liu, Zexiang and Huang, Zixuan and Letts, Adam and Li, Yangguang and Liang, Ding and Laforte, Christian and Jampani, Varun and Cao, Yan-Pei},
  journal={arXiv preprint arXiv:2403.02151},
  year={2024}
}

@inproceedings{hollein2024viewdiff,
  title={{ViewDiff}: {3D}-Consistent Image Generation with Text-to-Image Models},
  author={H{\"o}llein, Lukas and Bo{\v{z}}i{\v{c}}, Alja{\v{z}} and M{\"u}ller, Norman and Novotny, David and Tseng, Hung-Yu and Richardt, Christian and Zollh{\"o}fer, Michael and Nie{\ss}ner, Matthias},
  booktitle={Proc. CVPR},
  year={2024}
}

@inproceedings{szymanowicz2023splatter,
  title={Splatter image: Ultra-fast single-view {3D} reconstruction},
  author={Szymanowicz, Stanislaw and Rupprecht, Christian and Vedaldi, Andrea},
  booktitle={Proc. CVPR},
  year={2024}
}

@inproceedings{wang2023score,
  title={Score {Jacobian} chaining: Lifting pretrained 2d diffusion models for {3D} generation},
  author={Wang, Haochen and Du, Xiaodan and Li, Jiahao and Yeh, Raymond A and Shakhnarovich, Greg},
  booktitle={Proc. CVPR},
  year={2023}
}

@article{AdamW,
  title={Decoupled weight decay regularization},
  author={Loshchilov, I},
  journal={arXiv preprint arXiv:1711.05101},
  year={2017}
}

@article{yu2024viewcrafter,
  title={Viewcrafter: Taming video diffusion models for high-fidelity novel view synthesis},
  author={Yu, Wangbo and Xing, Jinbo and Yuan, Li and Hu, Wenbo and Li, Xiaoyu and Huang, Zhipeng and Gao, Xiangjun and Wong, Tien-Tsin and Shan, Ying and Tian, Yonghong},
  journal={arXiv preprint arXiv:2409.02048},
  year={2024}
}

@inproceedings{dust3r,
      title={DUSt3R: Geometric 3D Vision Made Easy}, 
      author={Shuzhe Wang and Vincent Leroy and Yohann Cabon and Boris Chidlovskii and Jerome Revaud},
      booktitle = {Proc. CVPR},
      year = {2024}
}

@article{li2024controllable,
  title={Controllable Text-to-{3D} Generation via Surface-Aligned {Gaussian} Splatting},
  author={Li, Zhiqi and Chen, Yiming and Zhao, Lingzhe and Liu, Peidong},
  journal={arXiv preprint arXiv:2403.09981},
  year={2024}
}

@article{he2024gvgen,
  title={{GVGEN}: Text-to-{3D} Generation with Volumetric Representation},
  author={He, Xianglong and Chen, Junyi and Peng, Sida and Huang, Di and Li, Yangguang and Huang, Xiaoshui and Yuan, Chun and Ouyang, Wanli and He, Tong},
  journal={arXiv preprint arXiv:2403.12957},
  year={2024}
}

@inproceedings{xu2024grm,
  title={{GRM}: Large {Gaussian} Reconstruction Model for Efficient {3D} Reconstruction and Generation},
  author={Xu, Yinghao and Shi, Zifan and Yifan, Wang and Chen, Hansheng and Yang, Ceyuan and Peng, Sida and Shen, Yujun and Wetzstein, Gordon},
  booktitle={Proc. ECCV},
  year={2024}
}

@article{ye2024dreamreward,
  title={{DreamReward}: Text-to-{3D} Generation with Human Preference},
  author={Ye, Junliang and Liu, Fangfu and Li, Qixiu and Wang, Zhengyi and Wang, Yikai and Wang, Xinzhou and Duan, Yueqi and Zhu, Jun},
  journal={arXiv preprint arXiv:2403.14613},
  year={2024}
}

@article{zhang2024compress3d,
  title={{Compress3D}: a Compressed Latent Space for {3D} Generation from a Single Image},
  author={Zhang, Bowen and Yang, Tianyu and Li, Yu and Zhang, Lei and Zhao, Xi},
  journal={arXiv preprint arXiv:2403.13524},
  year={2024}
}

@article{han2024vfusion3d,
  title={{VFusion3D}: Learning Scalable {3D} Generative Models from Video Diffusion Models},
  author={Han, Junlin and Kokkinos, Filippos and Torr, Philip},
  journal={arXiv preprint arXiv:2403.12034},
  year={2024}
}

@article{jiang2024brightdreamer,
  title={BrightDreamer: Generic {3D} {Gaussian} Generative Framework for Fast Text-to-{3D} Synthesis},
  author={Jiang, Lutao and Wang, Lin},
  journal={arXiv preprint arXiv:2403.11273},
  year={2024}
}

@inproceedings{kim2023neuralfield,
  title={{NeuralField-LDM}: Scene generation with hierarchical latent diffusion models},
  author={Kim, Seung Wook and Brown, Bradley and Yin, Kangxue and Kreis, Karsten and Schwarz, Katja and Li, Daiqing and Rombach, Robin and Torralba, Antonio and Fidler, Sanja},
  booktitle={Proc. CVPR},
  year={2023}
}

@inproceedings{xie2024latte3d,
  title = {{LATTE3D}: Large-scale Amortized Text-To-{enhanced3D} Synthesis},
  author = {Kevin Xie and Jonathan Lorraine and Tianshi Cao and Jun Gao and James Lucas and Antonio Torralba and Sanja Fidler and Xiaohui Zeng},
  booktitle = {Proc. ECCV},
  year = {2024},
}

@article{tang2024lgm,
  title={{LGM}: Large Multi-View Gaussian Model for High-Resolution 3D Content Creation},
  author={Tang, Jiaxiang and Chen, Zhaoxi and Chen, Xiaokang and Wang, Tengfei and Zeng, Gang and Liu, Ziwei},
  journal={Proc. ECCV},
  year={2024}
}

@inproceedings{szymanowicz2023viewset,
  title={Viewset diffusion:(0-) image-conditioned 3d generative models from 2d data},
  author={Szymanowicz, Stanislaw and Rupprecht, Christian and Vedaldi, Andrea},
  booktitle={Proc. ICCV},
  year={2023}
}

@article{shriram2024realmdreamer,
  title={Realmdreamer: Text-driven 3d scene generation with inpainting and depth diffusion},
  author={Shriram, Jaidev and Trevithick, Alex and Liu, Lingjie and Ramamoorthi, Ravi},
  journal={arXiv preprint arXiv:2404.07199},
  year={2024}
}

@inproceedings{hollein2023text2room,
  title={Text2room: Extracting textured 3d meshes from 2d text-to-image models},
  author={H{\"o}llein, Lukas and Cao, Ang and Owens, Andrew and Johnson, Justin and Nie{\ss}ner, Matthias},
  booktitle={Proc. ICCV},
  year={2023}
}

@inproceedings{sun2023dreamcraft3d,
  title={{DreamCraft3D}: Hierarchical {3D} generation with bootstrapped diffusion prior},
  author={Sun, Jingxiang and Zhang, Bo and Shao, Ruizhi and Wang, Lizhen and Liu, Wen and Xie, Zhenda and Liu, Yebin},
  booktitle={Proc. ICLR},
  year={2024}
}

@article{feng2024fdgaussian,
  title={{FDGaussian}: Fast {Gaussian} Splatting from Single Image via Geometric-aware Diffusion Model},
  author={Feng, Qijun and Xing, Zhen and Wu, Zuxuan and Jiang, Yu-Gang},
  journal={arXiv preprint arXiv:2403.10242},
  year={2024}
}

@article{liu2024isotropic3d,
  title={{Isotropic3D}: Image-to-{3D} Generation Based on a Single CLIP Embedding},
  author={Liu, Pengkun and Wang, Yikai and Sun, Fuchun and Li, Jiafang and Xiao, Hang and Xue, Hongxiang and Wang, Xinzhou},
  journal={arXiv preprint arXiv:2403.10395},
  year={2024}
}

@article{tang2024pixel,
  title={Pixel-Aligned Multi-View Generation with Depth Guided Decoder},
  author={Tang, Zhenggang and Zhuang, Peiye and Wang, Chaoyang and Siarohin, Aliaksandr and Kant, Yash and Schwing, Alexander and Tulyakov, Sergey and Lee, Hsin-Ying},
  journal={arXiv preprint arXiv:2408.14016},
  year={2024}
}

@article{szymanowicz2025bolt3d,
  title={Bolt3d: Generating 3d scenes in seconds},
  author={Szymanowicz, Stanislaw and Zhang, Jason Y and Srinivasan, Pratul and Gao, Ruiqi and Brussee, Arthur and Holynski, Aleksander and Martin-Brualla, Ricardo and Barron, Jonathan T and Henzler, Philipp},
  journal={arXiv preprint arXiv:2503.14445},
  year={2025}
}

@article{liang2024wonderland,
  title={Wonderland: Navigating 3D Scenes from a Single Image},
  author={Liang, Hanwen and Cao, Junli and Goel, Vidit and Qian, Guocheng and Korolev, Sergei and Terzopoulos, Demetri and Plataniotis, Konstantinos N and Tulyakov, Sergey and Ren, Jian},
  journal={Proc. CVPR},
  year={2025}
}

@article{agarwal2025cosmos,
  title={Cosmos world foundation model platform for physical ai},
  author={Team Cosmos},
  journal={arXiv preprint arXiv:2501.03575},
  year={2025}
}

@inproceedings{zhang2024gs,
  title={Gs-lrm: Large reconstruction model for 3d gaussian splatting},
  author={Zhang, Kai and Bi, Sai and Tan, Hao and Xiangli, Yuanbo and Zhao, Nanxuan and Sunkavalli, Kalyan and Xu, Zexiang},
  booktitle={Proc. ECCV},
  year={2024},
}

@article{bahmani2024ac3d,
  title={AC3D: Analyzing and Improving 3D Camera Control in Video Diffusion Transformers},
  author={Bahmani, Sherwin and Skorokhodov, Ivan and Qian, Guocheng and Siarohin, Aliaksandr and Menapace, Willi and Tagliasacchi, Andrea and Lindell, David B and Tulyakov, Sergey},
  journal={Proc. CVPR},
  year={2025}
}

@article{bahmani2024vd3d,
  title={Vd3d: Taming large video diffusion transformers for 3d camera control},
  author={Bahmani, Sherwin and Skorokhodov, Ivan and Siarohin, Aliaksandr and Menapace, Willi and Qian, Guocheng and Vasilkovsky, Michael and Lee, Hsin-Ying and Wang, Chaoyang and Zou, Jiaxu and Tagliasacchi, Andrea and others},
  journal={Proc. ICLR},
  year={2025}
}

@inproceedings{charatan2024pixelsplat,
  title={pixelsplat: 3d gaussian splats from image pairs for scalable generalizable 3d reconstruction},
  author={Charatan, David and Li, Sizhe Lester and Tagliasacchi, Andrea and Sitzmann, Vincent},
  booktitle={Proc. CVPR},
  year={2024}
}

@article{yu2025trajectorycrafter,
  title={Trajectorycrafter: Redirecting camera trajectory for monocular videos via diffusion models},
  author={YU, Mark and Hu, Wenbo and Xing, Jinbo and Shan, Ying},
  journal={Proc. ICCV},
  year={2025}
}

@article{liang2024btimer,
  title={Feed-Forward Bullet-Time Reconstruction of Dynamic Scenes from Monocular Videos},
  author={Liang, Hanxue and Ren, Jiawei and Mirzaei, Ashkan and Torralba, Antonio and Liu, Ziwei and Gilitschenski, Igor and Fidler, Sanja and Oztireli, Cengiz and Ling, Huan and Gojcic, Zan and Huang, Jiahui},
  journal={Proc. NeurIPS},
  year={2025}
  }

@article{schwarz2025generative,
  title={Generative Gaussian splatting: Generating 3D scenes with video diffusion priors},
  author={Schwarz, Katja and Mueller, Norman and Kontschieder, Peter},
  journal={arXiv preprint arXiv:2503.13272},
  year={2025}
}

@article{wang2025act,
  title={ACT-R: Adaptive Camera Trajectories for Single View 3D Reconstruction},
  author={Wang, Yizhi and Zhao, Mingrui and Mahdavi-Amiri, Ali and Zhang, Hao},
  journal={arXiv preprint arXiv:2505.08239},
  year={2025}
}

@inproceedings{kant2025pippo,
  title={Pippo: High-resolution multi-view humans from a single image},
  author={Kant, Yash and Weber, Ethan and Kim, Jin Kyu and Khirodkar, Rawal and Zhaoen, Su and Martinez, Julieta and Gilitschenski, Igor and Saito, Shunsuke and Bagautdinov, Timur},
  booktitle={Proc. CVPR},
  year={2025}
}

@article{lu2024infinicube,
  title={Infinicube: Unbounded and controllable dynamic 3d driving scene generation with world-guided video models},
  author={Lu, Yifan and Ren, Xuanchi and Yang, Jiawei and Shen, Tianchang and Wu, Zhangjie and Gao, Jun and Wang, Yue and Chen, Siheng and Chen, Mike and Fidler, Sanja and others},
  journal={arXiv preprint arXiv:2412.03934},
  year={2024}
}

@inproceedings{yuan2025generative,
  title={Generative photography: Scene-consistent camera control for realistic text-to-image synthesis},
  author={Yuan, Yu and Wang, Xijun and Sheng, Yichen and Chennuri, Prateek and Zhang, Xingguang and Chan, Stanley},
  booktitle={Proc. CVPR},
  year={2025}
}

@article{yang2025matrix,
  title={Matrix-3D: Omnidirectional Explorable 3D World Generation},
  author={Yang, Zhongqi and Ge, Wenhang and Li, Yuqi and Chen, Jiaqi and Li, Haoyuan and An, Mengyin and Kang, Fei and Xue, Hua and Xu, Baixin and Yin, Yuyang and others},
  journal={arXiv preprint arXiv:2508.08086},
  year={2025}
}

@article{zhang2025spatialcrafter,
  title={SpatialCrafter: Unleashing the Imagination of Video Diffusion Models for Scene Reconstruction from Limited Observations},
  author={Zhang, Songchun and Xu, Huiyao and Guo, Sitong and Xie, Zhongwei and Bao, Hujun and Xu, Weiwei and Zou, Changqing},
  journal={arXiv preprint arXiv:2505.11992},
  year={2025}
}

@article{kirschstein2025avat3r,
  title={Avat3r: Large Animatable Gaussian Reconstruction Model for High-fidelity 3D Head Avatars},
  author={Kirschstein, Tobias and Romero, Javier and Sevastopolsky, Artem and Nie{\ss}ner, Matthias and Saito, Shunsuke},
  journal={arXiv preprint arXiv:2502.20220},
  year={2025}
}

@article{schneider2025worldexplorer,
  title={WorldExplorer: Towards Generating Fully Navigable 3D Scenes},
  author={Schneider, Manuel-Andreas and H{\"o}llein, Lukas and Nie{\ss}ner, Matthias},
  journal={arXiv preprint arXiv:2506.01799},
  year={2025}
}

@article{ren2024scube,
  title={Scube: Instant large-scale scene reconstruction using voxsplats},
  author={Ren, Xuanchi and Lu, Yifan and Liang, Hanxue and Wu, Zhangjie and Ling, Huan and Chen, Mike and Fidler, Sanja and Williams, Francis and Huang, Jiahui},
  journal={Proc. NeurIPS},
  year={2024}
}

@article{xu20254dgt,
  title={4DGT: Learning a 4D Gaussian Transformer Using Real-World Monocular Videos},
  author={Xu, Zhen and Li, Zhengqin and Dong, Zhao and Zhou, Xiaowei and Newcombe, Richard and Lv, Zhaoyang},
  journal={arXiv preprint arXiv:2506.08015},
  year={2025}
}

@inproceedings{ren2024xcube,
  title={Xcube: Large-scale 3d generative modeling using sparse voxel hierarchies},
  author={Ren, Xuanchi and Huang, Jiahui and Zeng, Xiaohui and Museth, Ken and Fidler, Sanja and Williams, Francis},
  booktitle={Proc. CVPR},
  year={2024}
}

@article{li2025hunyuan,
  title={Hunyuan-gamecraft: High-dynamic interactive game video generation with hybrid history condition},
  author={Li, Jiaqi and Tang, Junshu and Xu, Zhiyong and Wu, Longhuang and Zhou, Yuan and Shao, Shuai and Yu, Tianbao and Cao, Zhiguo and Lu, Qinglin},
  journal={arXiv preprint arXiv:2506.17201},
  year={2025}
}

@article{tang2025hunyuan,
  title={Hunyuan-GameCraft-2: Instruction-following Interactive Game World Model},
  author={Tang, Junshu and Liu, Jiacheng and Li, Jiaqi and Wu, Longhuang and Yang, Haoyu and Zhao, Penghao and Gong, Siruis and Yuan, Xiang and Shao, Shuai and Lu, Qinglin},
  journal={arXiv preprint arXiv:2511.23429},
  year={2025}
}

@article{hong2025relic,
  title={Relic: Interactive video world model with long-horizon memory},
  author={Hong, Yicong and Mei, Yiqun and Ge, Chongjian and Xu, Yiran and Zhou, Yang and Bi, Sai and Hold-Geoffroy, Yannick and Roberts, Mike and Fisher, Matthew and Shechtman, Eli and others},
  journal={arXiv preprint arXiv:2512.04040},
  year={2025}
}

@article{zhang2025matrix,
  title={Matrix-game: Interactive world foundation model},
  author={Zhang, Yifan and Peng, Chunli and Wang, Boyang and Wang, Puyi and Zhu, Qingcheng and Kang, Fei and Jiang, Biao and Gao, Zedong and Li, Eric and Liu, Yang and others},
  journal={arXiv preprint arXiv:2506.18701},
  year={2025}
}

@article{he2025matrix,
  title={Matrix-game 2.0: An open-source real-time and streaming interactive world model},
  author={He, Xianglong and Peng, Chunli and Liu, Zexiang and Wang, Boyang and Zhang, Yifan and Cui, Qi and Kang, Fei and Jiang, Biao and An, Mengyin and Ren, Yangyang and others},
  journal={arXiv preprint arXiv:2508.13009},
  year={2025}
}

@article{xiao2025worldmem,
  title={Worldmem: Long-term consistent world simulation with memory},
  author={Xiao, Zeqi and Lan, Yushi and Zhou, Yifan and Ouyang, Wenqi and Yang, Shuai and Zeng, Yanhong and Pan, Xingang},
  journal={arXiv preprint arXiv:2504.12369},
  year={2025}
}

@article{zhao2025spatia,
  title={Spatia: Video Generation with Updatable Spatial Memory},
  author={Zhao, Jinjing and Wei, Fangyun and Liu, Zhening and Zhang, Hongyang and Xu, Chang and Lu, Yan},
  journal={arXiv preprint arXiv:2512.15716},
  year={2025}
}

@article{li2025magicworld,
  title={Magicworld: Interactive geometry-driven video world exploration},
  author={Li, Guangyuan and Zheng, Siming and Xu, Shuolin and Chen, Jinwei and Li, Bo and Hu, Xiaobin and Zhao, Lei and Jiang, Peng-Tao},
  journal={arXiv preprint arXiv:2511.18886},
  year={2025}
}

@inproceedings{zhang2025framepack,
    title={Frame Context Packing and Drift Prevention in Next-Frame-Prediction Video Diffusion Models},
    author={Lvmin Zhang and Shengqu Cai and Muyang Li and Gordon Wetzstein and Maneesh Agrawala},
    booktitle=NEURIPS,
    year={2025},
}

@article{po2025bagger,
  title={BAgger: Backwards Aggregation for Mitigating Drift in Autoregressive Video Diffusion Models},
  author={Po, Ryan and Chan, Eric Ryan and Chen, Changan and Wetzstein, Gordon},
  journal={arXiv preprint arXiv:2512.12080},
  year={2025}
}

@inproceedings{li2026flashworld,
  title={FlashWorld: High-quality 3D Scene Generation within Seconds},
  author={Li, Xinyang and Wang, Tengfei and Gu, Zixiao and Zhang, Shengchuan and Guo, Chunchao and Cao, Liujuan},
  booktitle=ICLR,
  year={2026}
}

@article{museth2013vdb,
  title={VDB: High-resolution sparse volumes with dynamic topology},
  author={Museth, Ken},
  journal={ACM Transactions on Graphics (TOG)},
  volume={32},
  number={3},
  pages={1--22},
  year={2013},
  publisher={ACM New York, NY, USA}
}

@article{huang2025selfforcing,
  title={Self Forcing: Bridging the Train-Test Gap in Autoregressive Video Diffusion},
  author={Huang, Xun and Li, Zhengqi and He, Guande and Zhou, Mingyuan and Shechtman, Eli},
  journal={arXiv preprint arXiv:2506.08009},
  year={2025}
}

@article{lipman2022flow,
  title={Flow matching for generative modeling},
  author={Lipman, Yaron and Chen, Ricky TQ and Ben-Hamu, Heli and Nickel, Maximilian and Le, Matt},
  journal={arXiv preprint arXiv:2210.02747},
  year={2022}
}

@article{gu2025long,
  title={Long-context autoregressive video modeling with next-frame prediction},
  author={Gu, Yuchao and Mao, Weijia and Shou, Mike Zheng},
  journal={arXiv preprint arXiv:2503.19325},
  year={2025}
}

@article{zhou2025learning,
  title={Learning 3d persistent embodied world models},
  author={Zhou, Siyuan and Du, Yilun and Yang, Yuncong and Han, Lei and Chen, Peihao and Yeung, Dit-Yan and Gan, Chuang},
  journal={arXiv preprint arXiv:2505.05495},
  year={2025}
}

@inproceedings{liu2025dynamem,
  title={Dynamem: Online dynamic spatio-semantic memory for open world mobile manipulation},
  author={Liu, Peiqi and Guo, Zhanqiu and Warke, Mohit and Chintala, Soumith and Paxton, Chris and Shafiullah, Nur Muhammad Mahi and Pinto, Lerrel},
  booktitle={ICRA},
  year={2025},
}

@article{wu2025geometry,
  title={Geometry forcing: Marrying video diffusion and 3d representation for consistent world modeling},
  author={Wu, Haoyu and Wu, Diankun and He, Tianyu and Guo, Junliang and Ye, Yang and Duan, Yueqi and Bian, Jiang},
  journal={arXiv preprint arXiv:2507.07982},
  year={2025}
}

@article{savov2025statespacediffuser,
  title={Statespacediffuser: Bringing long context to diffusion world models},
  author={Savov, Nedko and Kazemi, Naser and Zhang, Deheng and Paudel, Danda Pani and Wang, Xi and Van Gool, Luc},
  journal={arXiv preprint arXiv:2505.22246},
  year={2025}
}

@article{zhang2025test,
  title={Test-time training done right},
  author={Zhang, Tianyuan and Bi, Sai and Hong, Yicong and Zhang, Kai and Luan, Fujun and Yang, Songlin and Sunkavalli, Kalyan and Freeman, William T and Tan, Hao},
  journal={arXiv preprint arXiv:2505.23884},
  year={2025}
}

@inproceedings{dalal2025one,
  title={One-minute video generation with test-time training},
  author={Dalal, Karan and Koceja, Daniel and Xu, Jiarui and Zhao, Yue and Han, Shihao and Cheung, Ka Chun and Kautz, Jan and Choi, Yejin and Sun, Yu and Wang, Xiaolong},
  booktitle=CVPR,
  year={2025}
}

@article{hong2024slowfast,
  title={Slowfast-vgen: Slow-fast learning for action-driven long video generation},
  author={Hong, Yining and Liu, Beide and Wu, Maxine and Zhai, Yuanhao and Chang, Kai-Wei and Li, Linjie and Lin, Kevin and Lin, Chung-Ching and Wang, Jianfeng and Yang, Zhengyuan and others},
  journal={arXiv preprint arXiv:2410.23277},
  year={2024}
}

@article{seo2024genwarp,
  title={Genwarp: Single image to novel views with semantic-preserving generative warping},
  author={Seo, Junyoung and Fukuda, Kazumi and Shibuya, Takashi and Narihira, Takuya and Murata, Naoki and Hu, Shoukang and Lai, Chieh-Hsin and Kim, Seungryong and Mitsufuji, Yuki},
  journal=NEURIPS,
  year={2024}
}

@article{deutsch2026ppisp,
  title={PPISP: Physically-Plausible Compensation and Control of Photometric Variations in Radiance Field Reconstruction},
  author={Deutsch, Isaac and Mo{\"e}nne-Loccoz, Nicolas and State, Gavriel and Gojcic, Zan},
  journal={arXiv preprint arXiv:2601.18336},
  year={2026}
}

@article{williams2024fvdb,
  title={fvdb: A deep-learning framework for sparse, large scale, and high performance spatial intelligence},
  author={Williams, Francis and Huang, Jiahui and Swartz, Jonathan and Klar, Gergely and Thakkar, Vijay and Cong, Matthew and Ren, Xuanchi and Li, Ruilong and Fuji-Tsang, Clement and Fidler, Sanja and others},
  journal={ACM Transactions on Graphics (TOG)},
  volume={43},
  number={4},
  pages={1--15},
  year={2024},
  publisher={ACM New York, NY, USA}
}

@inproceedings{yin2024dmd,
  title={One-step diffusion with distribution matching distillation},
  author={Yin, Tianwei and Gharbi, Micha{\"e}l and Zhang, Richard and Shechtman, Eli and Durand, Fr{\'e}do and Freeman, William T and Park, Taesung},
  booktitle={CVPR},
  year={2024}
}

@article{hyworld2025,
  title={HY-World 1.5: A Systematic Framework for Interactive World Modeling with Real-Time Latency and Geometric Consistency},
  author={Team HunyuanWorld},
  journal={arXiv preprint},
  year={2025}
}

@inproceedings{zhao2023unipc,
  title={UniPC: A Unified Predictor-Corrector Framework for Fast Sampling of Diffusion Models},
  author={Zhao, Wenliang and Bai, Lujia and Rao, Yongming and Zhou, Jie and Lu, Jiwen},
  booktitle={NeurIPS},
  year={2023}
}

@inproceedings{ren2022look,
  title={Look Outside the Room: Synthesizing A Consistent Long-Term 3D Scene Video from A Single Image},
  author={Ren, Xuanchi and Wang, Xiaolong},
  booktitle={Proceedings of the IEEE/CVF Conference on Computer Vision and Pattern Recognition (CVPR)},
  year={2022}
}

@article{hoellein2026world,
      title={World Reconstruction From Inconsistent Views}, 
      author={Lukas H{\"o}llein and Matthias Nie{\ss}ner},
      year={2026},
      journal={arXiv preprint arXiv:2603.16736},
}

@article{shabanov2026free,
  title={Free-Range Gaussians: Non-Grid-Aligned Generative 3D Gaussian Reconstruction},
  author={Shabanov, Ahan and Hedman, Peter and Weber, Ethan and Li, Zhengqin and Rozumny, Denis and Lan, Gael Le and Dhingra, Naina and Luo, Lei and Vedaldi, Andrea and Richardt, Christian and others},
  journal={arXiv preprint arXiv:2604.04874},
  year={2026}
}

@inproceedings{gao2022get3d,
title={GET3D: A Generative Model of High Quality 3D Textured Shapes Learned from Images},
author={Jun Gao and Tianchang Shen and Zian Wang and Wenzheng Chen and Kangxue Yin and Daiqing Li and Or Litany and Zan Gojcic and Sanja Fidler},
booktitle={Proc. NeurIPS},
year={2022}
}
}
\end{document}